\documentclass[10pt,journal,compsoc]{IEEEtran}

\usepackage[nocompress]{cite}
\usepackage{url}
\usepackage{graphicx}
\usepackage{multirow}
\usepackage{enumitem}
\usepackage{bm}
\usepackage{amsmath}
\usepackage{amsfonts} 
\usepackage{xcolor,colortbl}
\usepackage{booktabs}
\usepackage[labelfont=bf,textfont=bf,skip=3pt,font=small]{caption}
\definecolor{Gray1}{rgb}{0.92,0.92,0.92}
\definecolor{Gray2}{rgb}{0.88,0.88,0.88}

\usepackage[bookmarks=false,colorlinks=true,linkcolor=black,citecolor=black,filecolor=black,urlcolor=black]{hyperref}

\begin{document}

\title{TransVG++: End-to-End Visual Grounding with Language Conditioned Vision Transformer}

\author{Jiajun Deng, 
        Zhengyuan Yang, 
        Daqing Liu, 
        Tianlang Chen, 
        Wengang Zhou,~\IEEEmembership{Senior Member,~IEEE,}\\
        Yanyong Zhang,~\IEEEmembership{Fellow,~IEEE,}
		Houqiang Li,~\IEEEmembership{Fellow,~IEEE,}
		Wanli Ouyang,~\IEEEmembership{Senior Member,~IEEE}
\IEEEcompsocitemizethanks{\IEEEcompsocthanksitem J. Deng, W. Zhou, Y. Zhang and H. Li are with University of Science and Technology of China. W. Zhou, Y. Zhang and H. Li are also with Institute of Artificial Intelligence, Hefei Comprehensive National Science Center. (email: dengjj@ustc.edu.cn, zhwg@ustc.edu.cn, yanyongz@ustc.edu.cn, lihq@ustc.edu.cn)
\IEEEcompsocthanksitem Z. Yang is with Microsoft. (email: zhengyuan.yang13@gmail.com)
\IEEEcompsocthanksitem D.Liu is with JD Explore Academy. (email: liudq.ustc@gmail.com)
\IEEEcompsocthanksitem T. Chen is with Amazon. (email: sunnychencool@gmail.com)
\IEEEcompsocthanksitem W. Ouyang is with the University of Sydney and Shanghai Artificial Intelligent Laboratory. (email: wanli.ouyang@sydney.edu.au)
\IEEEcompsocthanksitem Corresponding authors: W. Zhou and H. Li. 
}
}

\markboth{IEEE TRANS. ON PATTERN RECOGNITION AND MACHINE INTELLIGENCE, VOL. X, NO. X, June 2022}%
{Shell \MakeLowercase{\textit{et al.}}: Bare Advanced Demo of IEEEtran.cls for IEEE Computer Society Journals}

\IEEEtitleabstractindextext{%
\begin{abstract}

In this work, we explore neat yet effective Transformer-based frameworks for visual grounding. The previous methods generally address the core problem of visual grounding, \emph{i.e.}, multi-modal fusion and reasoning, with manually-designed mechanisms. Such heuristic designs are not only complicated but also make models easily overfit specific data distributions. To avoid this, we first propose TransVG, which establishes multi-modal correspondences by Transformers and localizes referred regions by directly regressing box coordinates. We empirically show that complicated fusion modules can be replaced by a simple stack of Transformer encoder layers with higher performance. However, the core fusion Transformer in TransVG is stand-alone against uni-modal encoders, and thus should be trained from scratch on limited visual grounding data, which makes it hard to be optimized and leads to sub-optimal performance. To this end, we further introduce TransVG++ to make two-fold improvements. For one thing, we upgrade our framework to a purely Transformer-based one by leveraging Vision Transformer (ViT) for vision feature encoding. For another, we devise Language Conditioned Vision Transformer that removes external fusion modules and reuses the uni-modal ViT for vision-language fusion at the intermediate layers. We conduct extensive experiments on five prevalent datasets, and report a series of state-of-the-art records.
\end{abstract}

}

\maketitle

\section{Introduction}
\label{sec:introduction}

\begin{figure*}
  \centering
  \includegraphics[width=0.99\linewidth]{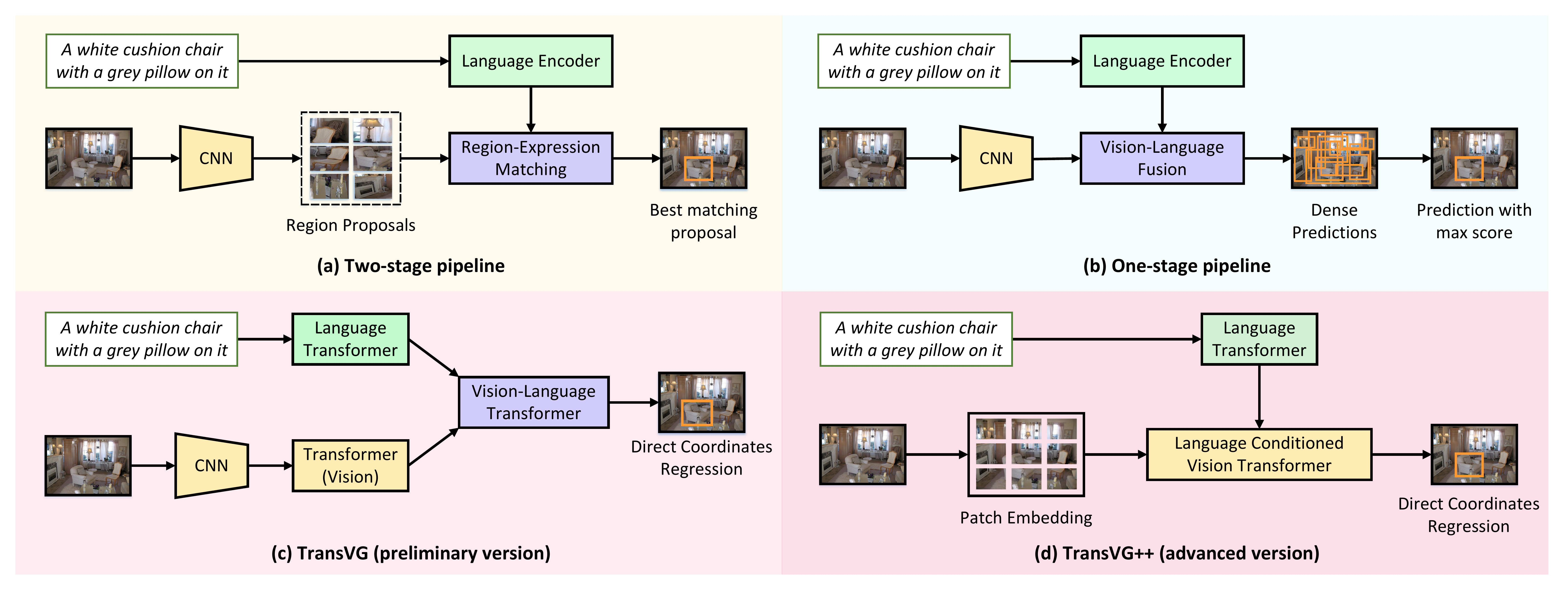}
  \caption{A comparison of (a) two-stage pipeline, (b) one-stage pipeline, (c) our preliminary TransVG framework, and (d) our advanced TransVG++ framework. TransVG performs intra- and inter-modality relation reasoning with a stack of Transformer layers in a homogeneous way, and grounds the referred region by directly regressing the box coordinates. TransVG++ takes further steps that upgrade the hybrid architecture composed of convolutional neural networks and Transformer networks to purely Transformer-based one and remove the stand-alone fusion Transformer by introducing Language Conditioned Vision Transformer to enable fusion at the intermediate layers of a vision feature encoder. Particularly, we use Transformer (vision) to represent a stack of Transformer encoder layers built on the top of a convolutional neural network (CNN), distinguished from Vision Transformer~\cite{dosovitskiy2020image}, a specific model in the literature.}
  \label{fig:intro}
\end{figure*}

\IEEEPARstart{V}{isual} grounding, which aims to localize a region referred to by a language expression in an image, is a core technology to bridge the natural language expression delivered by human beings and visual contents in the physical world.
The evolution of this eachnique is of great potential to promote vision-language understanding, and to provide an intelligent interface for human-machine interaction. Existing methods addressing this task generally follow two-stage or one-stage pipelines shown in 
Figure~\ref{fig:intro}. Specifically, two-stage approaches~\cite{nagaraja2016modeling,wang2019learning} first generate a set of region proposals, and then take visual grounding as a natural language object retrieval problem~\cite{hu2016natural,li2017deep} to find the best matching region given language expressions. Differently, one-stage approaches~\cite{chen2018real,liao2020real,yang2019fast} perform vision-language fusion at the output of the vision backbone network and the language model. Then, they make dense predictions with a sliding window over pre-defined anchor boxes, and keep the box with the maximum score as the final prediction.


Multi-modal fusion and reasoning is the primary problem in visual grounding. The early two-stage and one-stage methods address multi-modal fusion in a simple way. Concretely, the pioneer two-stage method, Embedding Net~\cite{wang2016learning}, measures the similarity between region embedding and expression embedding by cosine distance. The early one-stage approach, FAOA~\cite{yang2019fast}, encodes the language feature vector to each spatial position of vision feature maps by directly concatenating them. In general, these attempts are efficient, but lead to sub-optimal results, especially when it comes to complicated language expressions~\cite{liu2019learning,yang2020improving}. Following studies have proposed diverse architectures to ameliorate their performance.
Among two-stage methods, modular attention network~\cite{yu2018mattnet}, various scene graphs~\cite{wang2019neighbourhood,yang2019dynamic,yang2020graph}, and multi-modal tree~\cite{hong2019learning,liu2019learning} are designed to improve multi-modal relationships modeling.
The one-stage method~\cite{yang2020improving} also explores query decomposition and  proposes a multi-round fusion mechanism.

Despite their effectiveness, these sophisticated fusion or matching modules are built on pre-assumed dependencies of language expressions and visual scenes, making the models easily overfit specific scenarios, such as certain query lengths or objects relationships. Meanwhile, these mechanisms limit the plenitudinous interaction between vision and language contexts, which also hurts the performance of visual grounding algorithms. Besides, even though the target is to localize the referred region, most of the previous methods achieve this target in an indirect way. They generally define surrogate problems of language-guided candidates matching, selection, and refinement, following the common practice of image-text retrieval and object detection. Therefore, extra efforts have to be devoted to obtain candidates, including region proposals~\cite{yu2016modeling,mao2016generation,wang2019learning} and pre-defined anchor boxes~\cite{yang2019fast,yang2020improving}. Since these methods' predictions are made out of candidates, their performance is easily influenced by the step to generate such candidates and by the heuristics to assign targets to candidates. 

In this work, we explore an alternative approach in fusion module design, and re-formulate the prediction processing back to a simple regression problem. We first present TransVG, the preliminary version framework to address the problem of visual grounding with Transformers. We empirically show that the structurized fusion modules can be replaced by a simple stack of Transformer encoder layers.
The insight behind our design is that the basic component of Transformer, \emph{i.e.}, attention module, is ready to establish intra- and inter-modality correspondence for vision and language inputs, despite we neither suppose any language expression structure nor visual layout.
The pipeline of TransVG is illustrated in Figure~\ref{fig:intro}(c). We first feed the image and language expression into two sibling branches. The Transformer built on the top of a convolutional neural network and language Transformer built on word embeddings are applied in these two branches to model the global cues in vision and language domains, respectively. Then, the abstracted vision tokens and language tokens are fused together, and a vision-langauge Transformer is leveraged to perform cross-modal relation reasoning. Particularly, at the prediction step, TransVG directly outputs 4-dim coordinates of a bounding box to localize the referred region, instead of making predictions based on a set of candidates.

Although Transformers have been leveraged to establish multi-modal dependencies, TransVG shares the same meta-architecture with previous two-stage and one-stage methods, which include two independent uni-modal feature encoders, a stand-alone multi-modal fusion or matching module, and a prediction module. Within this meta-architecture, the parameters of vision encoder and language encoder can be initialized with well pre-trained models, while the primary fusion module, \emph{i.e.}, vision-language Transformer, is left to be trained from scratch with limited visual grounding data. As a result, the stand-alone fusion Transformer is hard to be optimized and leads to sub-optimal performance.

To this end, we further introduce an advanced version framework, namely TransVG++, to remove the stand-alone fusion Transformer, and instead performs vision-language fusion by re-using the vision feature encoder. As shown in Fig.\ref{fig:intro}(d), we devise Language Conditioned Vision Transformer (LViT) to play the role of both vision feature encoding and vision-language reasoning. Technically, LViT is obtained with minimal adaptation and extra parameters, by integrating information of language expressions into intermediate layers of the plain Vision Transformer (ViT)~\cite{dosovitskiy2020image}. Two novel alternative strategies, \emph{i.e.}, namely \textit{language prompter} and \textit{language adapter}, are proposed to convert uni-modal visual encoder layers of ViT into linguistic conditioned visual encoder layers of LViT. 
Built on a fully Transformer architecture and waiving the external fusion Transformer, TransVG++ achieves consistently better performance with even smaller model size and less computation costs when compared with the preliminary version.
Furthermore, piggybacking on the successful ViT series allows TransVG++ to be scaling up with minimal efforts. We empirically show that larger LViT models consistently lead to better TransVG++ variants, and expect the trend to hold for stronger Transformer based vision backbones.

We benchmark the proposed frameworks on five prevalent datasets, including RefCOCO~\cite{yu2016modeling}, RefCOCO+~\cite{yu2016modeling}, RefCOCOg~\cite{mao2016generation}, ReferItGame~\cite{kazemzadeh2014referitgame}, Flickr30K Entities~\cite{plummer2017flickr30k}.
Our preliminary framework respectively achieves 83.38\%, 59.24\%, 68.71\%, 70.73\% and 79.10\% accuracy on the test set of these five datasets, consistently outperforming the previous two-stage and one-stage approaches. Remarkably, TransVG++ further achieves 4.99\%, 8.04\%, 8.32\%, 3.97\% and 2.39\% absolute improvements over the preliminary one, setting a series of state-of-the-art records.

In summary, we make three-fold contributions:
\begin{itemize}
	\item We propose the first purely Transformer-based framework to address the problem of visual grounding, and re-formulate the prediction process to directly regress the box coordinates of referred regions.
	\item We present an elegant view of how to capture intra- and inter-modality context homogeneously with Transformers, and further investigate how to remove stand-alone fusion modules by integrating language information into intermediate layers of a vision encoder.
	\item We conduct extensive experiments to validate the merits of our method, and show significantly improved results on several prevalent benchmarks.
\end{itemize}

The preliminary version of this work is published in~\cite{deng2021transvg}. We have made significant improvements and extensions to our preliminary work. The major technical improvements can be concluded in two aspects. For the most important point, we remove the stand-alone fusion module, and enable the vision encoder to be re-used for multi-modal fusion at the intermediate layers. Besides, we upgrade the hybrid architecture composed of convolutional neural networks and Transformer networks to a purely Transformer-based one. To the best of our knowledge, this is the first fully Transformer-based framework in this field, without introducing any inductive bias. As demonstrated by experiemental results, the advanced framework outperforms the preliminary one by a large margin, and meanwhile achieves better efficiency on model size and computation costs. To facilitate further investigation, we build a benchmark of Transformer-based visual grounding frameworks at \url{https://github.com/djiajunustc/TransVG}, and will make our codes and models available.

\section{Related Work}

\subsection{Visual Grounding}
Visual grounding aims to ground a natural language description onto the referred region in an image. Most existing approaches consist of a vision feature encoder, a language feature encoder, and an external vision-language fusion module. According to the differences in the fusion module, current visual grounding methods can be broadly categorized into two directions, \emph{i.e.}, two-stage methods~\cite{hong2019learning,hu2017modeling,liu2019learning,wang2019learning,wang2019neighbourhood,yang2019dynamic,yu2018mattnet,zhang2018grounding,zhuang2018parallel} and one-stage methods~\cite{chen2018real,liao2020real,sadhu2019zero,yang2020improving,yang2019fast}. 
Two-stage methods match the language feature to the vision content at the region level, thus requiring the vision encoder to first generate a set of region proposals. One-stage methods densely perform multi-modal feature fusion at all spatial locations, waiving the requirements of region proposals. 
In the following, we deliver a literature review on both of them.


\noindent\textbf{Two-stage Methods.}
Inspired by the success of region based object detectors~\cite{girshick2014rich, girshick2015fast, ren2016faster}, two-stage methods are characterized by first generating region proposals and then selecting the best matching one corresponding to the language expression. At the first stage, region proposals can be either generated with external modules based on super-pixels grouping~\cite{uijlings2013selective,plummerCITE2018,wang2019learning}, or predicted with pre-trained object detectors~\cite{anderson2018bottom,yu2018mattnet,zhang2018grounding,liu2019learning}. 
The main efforts of approaches in this direction are devoted to the second stage, addressing visual grounding as text-region matching. 
The pioneer studies~\cite{mao2016generation,wang2016learning,nagaraja2016modeling} obtain good results by optimizing the feature embedding networks with maximum-margin ranking loss to maximize the similarity between the positive object-query pairs. The following work DBNet~\cite{zhang2017discriminative} and Similarity Net~\cite{wang2019learning} show that the similarity and dissimilarity of text-region pairs can be predicted by directly performing binary classification. 
The comprehensive region proposals provided by the vision encoder make it easy for two-stage methods to reason the relationships among objects. MMI~\cite{yu2016modeling} first proposes to incorporate visual comparison based context into referring expression models, which demonstrates the significance of involving out-of-object information for visual grounding. MattNet~\cite{yu2018mattnet} introduces the modular design and improves the grounding accuracy by better modeling the subject, location, and relation-related language description. Recent studies further improve the two-stage methods by better modeling the object relationships~\cite{liu2019learning,wang2019neighbourhood,yang2019dynamic,yang2020relationship,yang2020graph}, enforcing correspondence learning~\cite{liu2019improving}, or making use of phrase co-occurrences~\cite{bajaj2019g3raphground,chen2017query,dogan2019neural}.

\noindent \textbf{One-stage Methods.} 
One-stage approaches~\cite{chen2018real,liao2020real,sadhu2019zero,yang2020improving,yang2019fast} get rid of the region proposals and fuse the linguistic context with visual features densely at each spatial position of feature maps. The fused image-text feature maps are then leveraged to perform bounding box prediction in a sliding-window manner. The early work FAOA~\cite{yang2019fast} encodes the language expression into a language vector, and fuses the language vector into the YOLOv3 detector~\cite{redmon2018yolov3} to ground the referred region. RCCF~\cite{liao2020real} formulates the visual grounding problem as a correlation filtering process~\cite{bolme2010visual,henriques2014high}, and picks the peak value of the correlation heatmap as the center of target objects. The recent work ReSC~\cite{yang2020improving}  devises a recursive sub-query construction module to address the limitations of FAOA~\cite{yang2019fast} on grounding complex queries by multi-round fusion. LSPN~\cite{yang2020propagating} further improves the reasoning ability of the one-stage grounding methods. 

\noindent \textbf{Fusion in the Vision Encoder.}
Abstracting away the distinctions in the fusion module, one- and two-stage visual grounding approaches share the same meta-framework, \emph{i.e.}, single-modal feature encoders followed by an external multi-modal feature fusion or matching module. In this study, we alternatively explore how to remove the stand-alone fusion module, and re-use the vision feature encoder (\emph{i.e.}, a plain ViT~\cite{dosovitskiy2020image} in our TransVG++) to perform vision-language fusion. We vision two major advantages of the proposed novel paradigm. First, removing the explicit fusion layers makes the framework efficient and light-weighted, thus also performing better with the same amount of computations. Second, the unified vision encoder design facilitates the model to benefit from the advances in computer vision studies, such as stronger image classification and object detection models pretrained on large scale datasets.

\begin{figure*}
	\centering
	\includegraphics[width=0.95\linewidth]{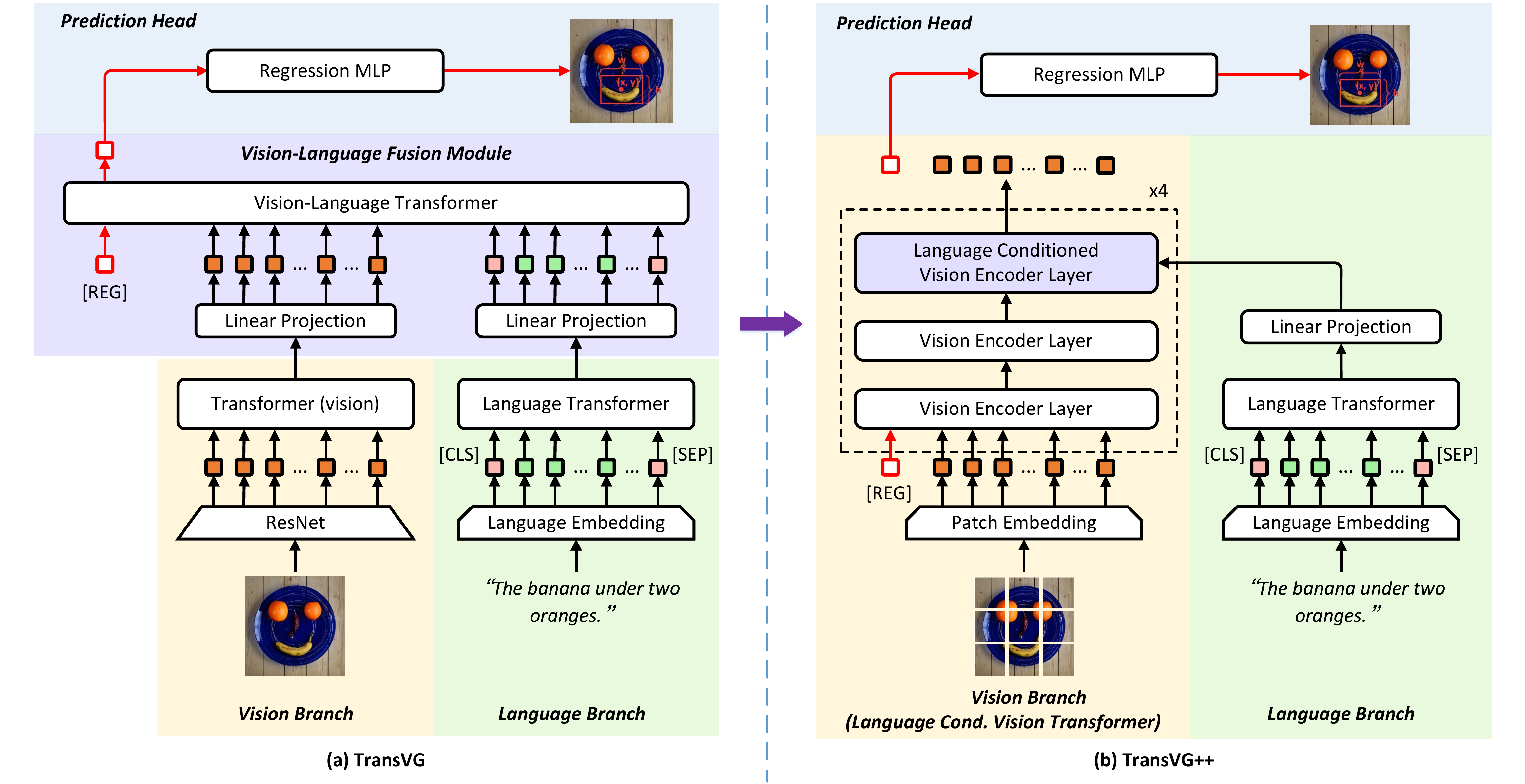}
	\caption{An overview of our proposed TransVG and TransVG++ frameworks. TransVG includes four components: a vision branch, a language branch, a vision-language fusion module and a prediction head. Vision tokens, language tokens and a learnable \texttt{[REG]} token are put together as the inputs of vision-language Transformer for multi-modal reasoning. The output state of \texttt{[REG]} token is fed into the prediction head for box cooridnates regression. To extend the preliminary framework, 
	TransVG++ removes the stand-alone fusion module, and introduce Langauge Conditioned Vision Transformer, which enables vision-langauge fusion at the intermediate layers of a vision feature encoder.
	}
	\label{fig:frameworks}
  \end{figure*}

\subsection{Transformer}
Transformer is first introduced in~\cite{vaswani2017attention} to tackle the problem of neural machine translation (NMT). Compared with recurrent units in RNNs~\cite{mikolov2010recurrent,tai2015improved} and LSTMs~\cite{hochreiter1997long}, the core component of Transformer, \emph{i.e.}, attention modules, show remarkable superiority in long-term sequence modeling. 
Transformer attracts increasing interests in the computer vision community, and has made unnegligible impacts on vision and vision-language tasks.

\noindent \textbf{Transformers in Vision Tasks.}
Inspired by the great success in NMT, a series of  Transformers~\cite{carion2020end,chen2020generative,dosovitskiy2020image,yang2020learning,zeng2020learning,touvron2021training,liu2021swin,meng2021conditional} applied to vision tasks have been proposed. The infusive work DETR~\cite{carion2020end} formulates object detection as a set prediction problem. It introduces a set of learnable object queries, reasons global context and object relations with attention mechanism, and outputs the final set of instance predictions. ViT~\cite{dosovitskiy2020image} and DeiT~\cite{touvron2021training} show that a pure Transformer can achieve excellent accuracy for image classification. The following work Swin Transformer~\cite{liu2021swin} re-introduces inductive bias into vision Transformers by devising shifted-window attention. The recent works UViT~\cite{chen2021simple} and ViTDet~\cite{li2022exploring} demonstrate the non-hierarchical ViT can be directly applied as the backbone network of object detectors. 
Our TransVG++ also explores the usage of plain ViT in downstream tasks. 
Distinguishly, we investigate how to adapt the uni-modal ViT for the multimodal visual grounding task, while best preserving its power and architecture for vision feature extraction.

\noindent \textbf{Transformer in Vision-Language Tasks.} 
Motivated by the powerful pre-trained model of BERT~\cite{devlin2018bert}, some researchers start to investigate visual-linguistic pre-training (VLP) ~\cite{chen2020uniter,li2020oscar,lu2019vilbert,su2019vl,yang2020tap} to jointly represent images and texts. In general, the early works take region proposals and text as inputs, and devise several transformer encoder layers for joint representation learning. Plenty of pre-training tasks are introduced, including image-text matching (ITM), word-region alignment (WRA), masked language modeling (MLM), masked region modeling (MRM), \emph{etc.} Some recent works~\cite{huang2020pixel,kim2021vilt,li2021align,wang2021simvlm} further improve the model to make use of raw image patches. Besides following the pretext tasks of BERT, MDETR~\cite{kamath2021mdetr} explores using object detection as the pretext task to help improve the performance of some downstream vision-language tasks.

Despite with similar base units, the goal of VLP is to learn a generalizable vision-language representation with large-scale data. In contrast, we focus on exploring novel Transformer-based visual grounding frameworks. Besides, ViT has has also shown its power in other vision-language tasks, while mainly as the backbone of vision encoder~\cite{li2021align,wang2021simvlm}. They still build stand-alone blocks for multi-modal fusion. We alternatively explore having the plain ViT do more work and removing the stand-alone fusion layers.

\section{Our Approach}

In this work, we present two versions of novel architecture for visual grounding by leveraging the glamorous Transformer neural networks. An overview of our proposed frameworks are illustrated in Fig.~\ref{fig:frameworks}. 
The preliminary version framework, namely TransVG, performs vision-language feature fusion with a simply stack of Transformer encoder layers, getting rid of manually designed mechanisms. 
The advanced version framework, \emph{i.e.}, TransVG++, is featured by removing the stand-alone fusion module and introducing Language Conditioned Vision Transformer (LViT), which play the roles of both vision feature encoding and multi-modal fusion. Particulary, LViT is converted from the uni-modal ViT~\cite{dosovitskiy2020image} model with minimal adaptation and extra costs.
Moreover, different from previous methods that rely on candidate boxes selection or refinement, both of our proposed frameworks introduce a learnable \texttt{[REG]} token, and localize the referred region by directly regressing the box coordiantes with the output state of \texttt{[REG]} token. 

In the Sec.~\ref{sec3:bg}, we first review the background knowledge of Transformer networks. Then, we elaborate our model design of the preliminary version (Sec.~\ref{sec3:transvg}) and advanced version (Sec.~\ref{sec3:transvg++}) frameworks. Finally, in Sec.~\ref{sec3:training_objectives}, we introduce the training objectives of our frameworks.

\begin{figure}
	\centering
	\includegraphics[width=0.95\linewidth]{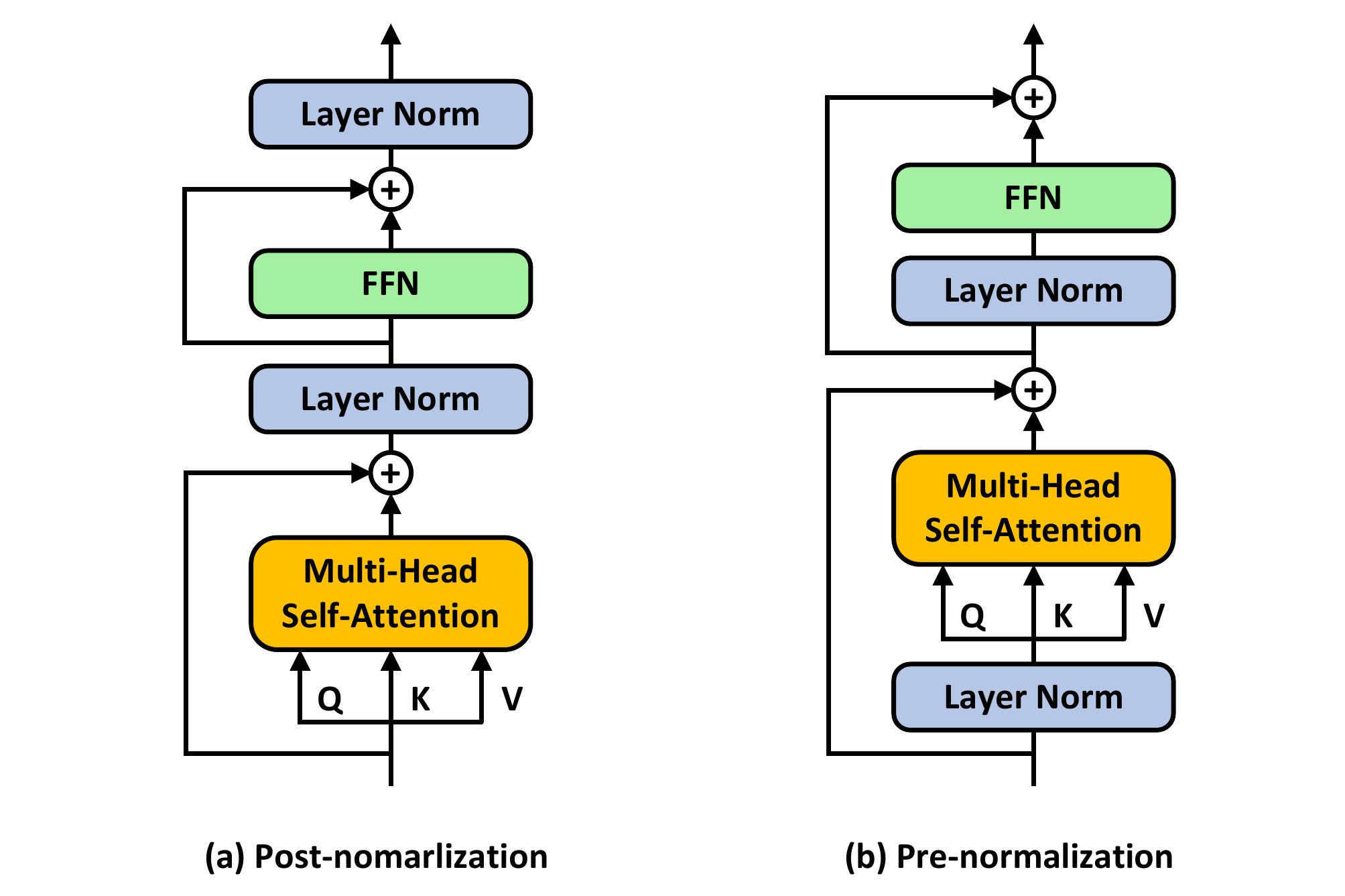}
	\caption{An illustration of two varieties of Transformer encoder layers, including (a) a post-normalization encoder layer and (b) a pre-normalization encoder layer.}
	\label{fig:encoder_layer}
  \end{figure}

\subsection{Background: Transformer} \label{sec3:bg}
Before detailing our proposed frameworks, we first deliver a background overview of attention modules and encoder layers of Transformer neural networks~\cite{vaswani2017attention}.

\noindent \textbf{Attention Module.}
The core component of Transformer encoder layer is the attention module. Here, we take single-head attention as an example for explanation. The input of single-head attention module is a sequence of feature tokens. Given the query tokens $\bm{x}^q$ and support tokens $\bm{x}^s$, three independent fully connected (FC) layers are applied on them to generate the query embedding $\bm{f}^Q$, key embedding $\bm{f}^K$ and value embedding $\bm{f}^V$ as follows:
\begin{align}
	\bm{f}^Q = \text{FC}(\bm{x}^q),\quad \bm{f}^K = \text{FC}(\bm{x}^s),\quad \bm{f}^V = \text{FC}(\bm{x}^s).
\end{align}
When support token set is the same as query token set, the attention module is named as a self-attention module. Otherwise, it is named as a cross-attention module. After obtaining query, key and value embedding, the output of a single-head attention layer is computed as:
\begin{equation}
	\label{func:attn}
	\text{Attn}(\bm{f}^Q,\bm{f}^K,\bm{f}^V) = \text{FC}(\text{Softmax}(\frac{\bm{f}^Q\bm{f}^K}{\sqrt{{d}^K}})\cdot\bm{f}^V),
\end{equation}
where $d^K$ is the channel dimension of $\bm{f}^K$, $\text{Softmax}(\cdot)$ is the softmax function applied across the key/value embedding.

\noindent \textbf{Encoder Layer.} 
In Fig.~\ref{fig:encoder_layer}, we illustrate two kinds of Transformer encoder layers, distinguished by the position to perform layer normalization (LN)~\cite{ba2016layer}. Concretely, a Transformer encoder layer has two main sub-layers, \emph{i.e.}, a multi-head self-attention (MHSA) layer and a feed forward network (FFN). Multi-head attention is a variant of single-head attention by splitting embedding channels into multiple groups. FFN is an multi-layer perceptron (MLP) composed of FC layers and non-linear activation layers. 

In Transformer encoder layers, each sub-layer is put into a residual structure. 
Let us denote the input as $\bm{x}_n$, a post-normalization Transformer encoder layer computes:
\begin{align}
	\bm{x}'_n &= \text{LN}(\bm{x}_n + \mathcal{F}_\text{MHSA}(\bm{x}_n)), \label{eq:post_msa} \\
	\bm{x}_{n+1} &= \text{LN}(\bm{x}'_n + \mathcal{F}_\text{FFN}(\bm{x}'_n)), \label{eq:post_FFN}
\end{align}
while the computation procedure in a pre-normalization Transformer encoder layer is:
\begin{align}
	\bm{x}'_n &= \bm{x}_n + \mathcal{F}_\text{MHSA}(\text{LN}(\bm{x}_n)), \label{eq:pre_msa} \\
	\bm{x}_{n+1} &= \bm{x}'_n + \mathcal{F}_\text{FFN}(\text{LN}(\bm{x}'_n)). \label{eq:pre_FFN}
\end{align}

\subsection{Preliminary Version: TransVG Framework} \label{sec3:transvg}
In this subsection, we present TransVG, the preliminary version framework based on a stack of Transformer encoder layers with direct box coordinates prediction. As shown in Figure~\ref{fig:frameworks}(a), given an image and a language expression as inputs, we first separate them into two sibling branches, \emph{i.e.}, a vision branch and a language branch, to generate corresponding feature embedding. Then, we consturct the inputs of vision-language fusion module by putting the vision and langauge feature embedding together, and append a learnable token (\emph{i.e.}, \texttt{[REG] token}).
The vision-language Transformer homogeneously embeds the input tokens from different modalities into a common semantic space by modeling intra- and inter-modality context with the self-attention layers. Finally, the output state of \texttt{[REG]} token is leveraged to directly predict 4-dim coordinates of a referred region in the prediction head.

\noindent \textbf{Vision Branch.}
The vision branch includes a convolutional network and a following Transformer encoder. We exploit the commonly used ResNet~\cite{he2016deep} as the convolutional backbone network, and build the Transformer encoder with a stack of 6 post-normalization Transformer encoder layers. In each encoder layer, There are 8 heads in the MHSA module, and 2 FC layers followed by ReLU activation layers in the FFN. The output channel dimensions of these 2 FC layers in the FFN are 2048 and 256, respectively.

Given an image $\bm{z}_\text{0}\in\mathbb{R}^{3\times H_0 \times W_0}$ as the input, we exploit the backbone network to generate a 2D feature map $\bm{z} \in\mathbb{R}^{C\times H \times W}$. Typically, the channel dimension $C$ is $2048$, and the width and height of $\bm{z}$ are $\frac{1}{32}$ of the original image size ($H=\frac{H_0}{32}$, $W=\frac{W_0}{32}$). Then, we leverage a $1\times1$ convolutional layer to reduce the channel dimension of $\bm{z}$ to $C_v=256 $ and obtain $\bm{z}' \in \mathbb{R}^{C_v \times H \times W}$. Since the input of a Transformer encoder layer is supposed to be a sequence of 1D vectors, we further flatten $\bm{z}'$ into $\bm{z}_v \in \mathbb{R}^{C_v\times N_v}$, where $N_v=H\times W$ is the number of input vision tokens. To make the visual Transformer sensitive to the original 2D positions of input tokens, we follow~\cite{carion2020end,parmar2018image} to utilize sine position embedding as the supplementary of visual feature. Concretely, the position encodings are added with the query and key embedding at each Transformer encoder layer. The visual Transformer conducts global context reasoning in parallel, and outputs visual embedding $\bm{f}_v$, which shares the same shape as input $\bm{z}_v$.

\noindent \textbf{Language Branch.}
The language branch is a sibling to the vision branch, and it includes a language embedding layer and a language Transformer. To make the best of pre-trained language models, the architecture of this branch follows BERT$_\text{BASE}$~\cite{devlin2018bert}. Typically, there are 12 pre-normalization Transformer encoder layers. The output channel dimension of language Transformer is $C_l=768$.

Given a language expression, we first represent each word ID as a one-hot vector. Then, in the language embedding layer, we convert each one-hot vector into a language token by looking up the token table. We follow the common practice in NMT~\cite{dehghani2018universal,devlin2018bert,raffel2019exploring,vaswani2017attention} to append a \texttt{[CLS]} token and a \texttt{[SEP]} token at the beginning and end positions of language tokens. Different from the sine position embedding leveraged in the vision branch, we make use of learnable position embedding in the language branch, and directly add them to language tokens. After that, we take these language tokens as inputs of linguistic Transformer to obtain the output language feature embedding $\bm{f}_l \in \mathbb{R}^{C_l \times N_l}$, where $N_l$ is the number of language tokens. 

\noindent \textbf{Vision-Language Fusion Module.}
As the core component in our model to fuse multi-modal information, the architecture of the vision-language fusion module (abbreviated as V-L module) is extremely simple and elegant. Specifically, the V-L module includes two linear projection layers (one for each modality) and a vision-language Transformer (also with a stack of 6 post-normalization Transformer encoder layers).

Given vision tokens $\bm{f}_v\in\mathbb{R}^{256\times N_v}$ out of the vision branch and language tokens $\bm{f}_l\in\mathbb{R}^{768\times N_l}$ out of the language branch, we apply a linear projection layer on each of them to project them into embedding with the same channel dimension. We denote the projected visual embedding and linguistic embedding as $\bm{g}_v\in{\mathbb{R}^{C_p\times N_v}}$ and $\bm{g}_l\in{\mathbb{R}^{C_p\times N_l}}$, where the projected feature dimension $C_p=256$. Then, we pre-append a learnable embedding (namely a \texttt{[REG]} token) to $\bm{g}_v$ and $\bm{g}_l$, and formulate the joint input tokens of the visual-linguistic Transformer as:
\begin{equation}
	\label{eq:token}
	\bm{x}_0 = [g_r,\ \underbrace{g_v^1,\ g_v^2,\ \cdots,\ g_v^{N_v}}_{\text{vision tokens}~\bm{g}_v},\ \overbrace{g_l^1,\ g_l^2,\ \cdots,\ g_l^{N_l}}^{\text{language tokens}~\bm{g}_l} ],
\end{equation}
where $g_r\in\mathbb{R}^{C_p \times1}$ represents the \texttt{[REG]} token. The \texttt{[REG]} token is randomly initialized at the beginning of the training stage and optimized with the whole model. After obtaining the input $\bm{x}_0\in\mathbb{R}^{C_p\times(1+N_v+N_l)}$ in the joint embedding space as described above, we apply the vision-language Transformer to embed $\bm{x}_0$ into a common semantic space by performing intra- and inter-modality relation reasoning in a homogeneous way. To retain the positional and modal information, we add learnable position embedding to the input of each Transformer encoder layer.

Thanks to the attention mechanism, the correspondence can be freely established between each pair of tokens from the joint entities, regardless of their modality. For example, a vision token can attend to a vision token, and it can also freely attend to a language token. Typically, the output state of the \texttt{[REG]} token develops a consolidated representation enriched by both vision and language context, and is further leveraged for box coordinates prediction.

\noindent \textbf{Prediction Head.}
We leverage the output state of \texttt{[REG]} token from the V-L module as the input of our prediction head. To perform box coordinates prediction, we apply a regression block to the \texttt{[REG]} token. The regression block is implemented by an MLP with two ReLU activated hidden layers and a linear output layer. The output of the prediction head is the 4-dim box coordinates.

\begin{figure}
	\centering
	\includegraphics[width=1.0\linewidth]{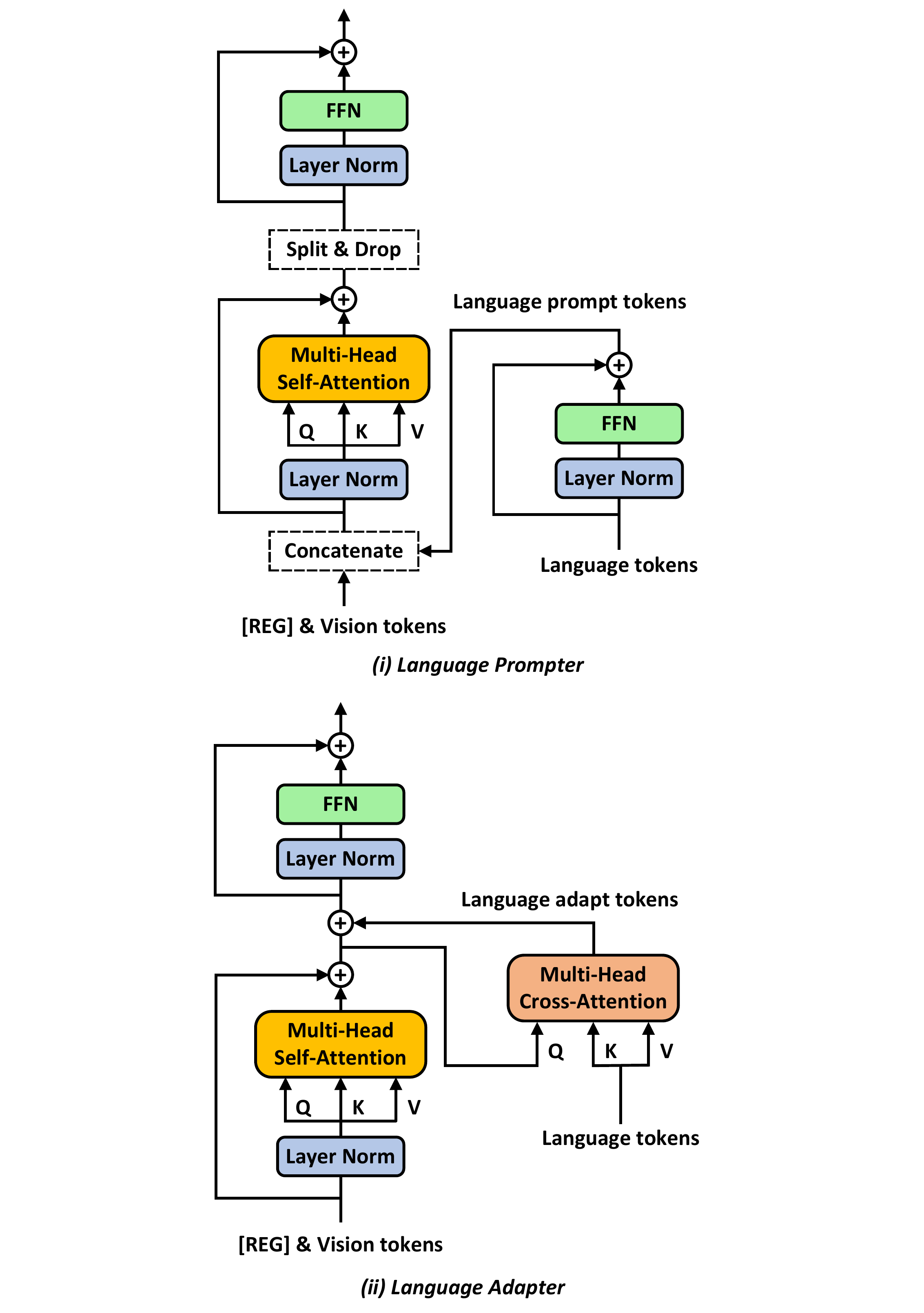}
	\caption{An illustration of our proposed language conditioned vision encoder layers with (i) a language prompter and (ii) a language adapter.
	}
	\label{fig:language_modulation}
  \end{figure}

\subsection{Advanced Version: TransVG++ Framework} \label{sec3:transvg++}

To further drive the evolution of Transformer-based visual grounding, we introduce TransVG++, the advanced version framework with fully Transformer-based architecture. We depict the overview of TransVG++ framework in Fig.~\ref{fig:frameworks}(b). As illustrated, the main difference between TransVG++ and TransVG is that we remove the stand-alone vision-language fusion module, and capitalize on a novel vision feature encoder, namely Language Contioned Vision Transformer (LViT). On the one hand, LViT upgrades the hybrid vision feature encoder composed of ResNet and 6 following Transfomer encoder layers, and on the other hand LViT enable multi-modal fusion at its intermediate layers. Besides, the vision input of our TransVG++ is no longer the whole images, but equally divided raw image patches.

Particularly, LViT is modified from a uni-modal ViT~\cite{dosovitskiy2020image} model with few efforts and negligible extra computation costs. Specifically, a standard ViT model is composed of 12 pre-normalization Transformer encoder layers. We denote them as vision encoder layers, since they are only leveraged to extract vision feature embedding, and equally divide them into 4 groups. Our core modification is that we convert the last vision encoder layer of each group into a language conditioned vision encoder layer, which integrates the language feature of referring expressions into vision tokens. Motivated by the success of prompt tuning~\cite{liu2021gpt} and network adapter~\cite{houlsby2019parameter,pfeiffer2020adapterfusion} in transfer pre-trained models to downstream tasks, we propose two simple yet effective strategies for language feature integration: (i) \textit{language prompter}, and (ii) \textit{language adapter}.

\textbf{Language Prompter}. We illustrates the architecture of a language conditioned vision encoder layer with language prompter in Fig.~\ref{fig:language_modulation}(i). In this strategy, we first feed the language tokens into a feed forward network (FFN) to generate language prompt tokens. After that, language prompt tokens are concatenated with the \texttt{[REG]} token and vision tokens to construct the input of a multi-head self-attention (MHSA) module. The tokens out of MHSA module is split into two groups, \emph{i.e.} one group for vision tokens and \texttt{[REG]} token, and another group for prompt tokens. The prompt tokens are dropped, while the other group is fed into the following FFN to obtain the input of next layers. Here, the parameters of the FFN to generate prompt tokens are randomly initialized, and those of other compnents make use of parameters from the corresponding original vision encoder layer of a pre-trained ViT backbone network.

\textbf{Language Adapter}.
The architecture of a language conditioned vision encoder layer with language adapter is illustrated in Fig.~\ref{fig:language_modulation}(ii). In this strategy, we inject the language expression information by adding an extra multi-head cross-attention (MHCA) module between the MHSA module and FFN of the original vision encoder layer. The query embedding of this MHCA module is the output of MHSA module, and both the key and value embedding are from the language tokens. By this way, the outputs of MHCA module, namely language adapt tokens, are the aggregation of language features with adaptive weights against each vision token. The language adapt tokens are then elementwisely added to the vision tokens for vision-language feature fusion. Similar to the language prompter, the parameters of the MHCA module are randomly initialized, and those of other parts are initialized with corresponding parameters from a pre-trained ViT backbone network.

By capitalizing on either a language prompter or a language adapter, the vision encoder layer can be converted to language conditioned vision encoder layers, and thus enables it to be re-used for absorbing text expression information from language tokens. The proposed LViT maintains the original architecture of a ViT model, so that we can naturally take advantage of pre-trained ViT model to ease the optimization for the whole framework. Particularly, both of these two strategies introduce negligible model parameters and computation costs compared with the whole model, while achieving infusive performance improvements against our preliminary version with stand-alone fusion Transformers. In our experiments, we conduct extensive ablative studies and analyses on these two strategies, and empirically show that using language adapter can achieve a slightly better accuracy among these two strategies.

\subsection{Training Objective} \label{sec3:training_objectives}
Unlike many previous methods that ground referred objects based on a set of candidates (\emph{i.e.}, region proposals in two-stage methods and anchor boxes in one-stage methods), we directly predict a 4-dim vector as the coordinates of the box to be grounded. This simplifies the process of target assignment and positive/negative examples mining at the training stage, but it also involves the scale problem. Specifically, the widely used smooth L1 loss tends to be a large number when we try to predict a large box, while tends to be small when we try to predict a small one, even if their predictions have similar relative errors. 

To address this problem, we normalize the coordinates of the ground-truth box by the scale of the image, and involve the generalized IoU loss~\cite{rezatofighi2019generalized} (GIoU loss), which is not affected by the scales. Let us denote the prediction as $\bm{b}=(x, y, w, h)$, and the normalized ground-truth box as $\hat{\bm{b}}=(\hat{x},\hat{y},\hat{w},\hat{h})$. The training objective of our method is:
\begin{equation}
	\label{func:loss}
	\mathcal{L} = \mathcal{L}_{\text{smooth-l1}}(\bm{b},\hat{\bm{b}}) + \mathcal{L}_{\text{giou}}(\bm{b},\hat{\bm{b}}),
\end{equation}
where $\mathcal{L}_{\text{smooth-l1}}(\cdot)$ and $\mathcal{L}_{\text{giou}}(\cdot)$ are the smooth L1 loss and GIoU loss, respectively.

\section{Experiments}
In this section, we present extensive experiments to validate the merits of our proposed TransVG/TransVG++ framework. In Sec.~\ref{sec4:datasets} and Sec.~\ref{sec4:impl}, we first give a brief introduction to the datasets and experimental setup. In Sec~\ref{sec4:analysis}, we conduct extensive ablative experiments to investigate the effectiveness of each component in our preliminary and advanced frameworks. Then, in Sec.~\ref{sec4:overall_comp}, we present the main results of our models, and compare them with other state-of-the-art methods. After that, in Sec.~\ref{sec4:discussion}, we discuss the model size and computation costs of TransVG++, and point out future directions to make it more applicable on resource-limited devices. Finally, in Sec.~\ref{sec4:examples}, we show some qualitative results to help validation and analysis.

\subsection{Datasets and Evaluation Metric}\label{sec4:datasets}
We conduct experiments on five prevalent datasets and follow the standard protocol to evaluate our framework in terms of accuracy. Specifically, only when the Jaccard overlap between the predicted region and the ground-truth region is above 0.5, the prediction is regarded as a correct one. These five datasets are detailed as follows:

\textit{RefCOCO}~\cite{yu2016modeling} includes 19,994 images with 50,000 referred objects. In each image, there are multiple instances belonging to the same categories. Each referred instance has more than one referring expression, and there are 142,210 referring expressions in total. The samples in RefCOCO are officially split into a train set with 120,624 expressions, a validation set with 10,834 expressions, a testA set with 5,657 expressions and a testB set with 5,095 expressions. 

\textit{RefCOCO+}~\cite{yu2016modeling} contains 19,992 images with 49,856 referred objects and 141,564 referring expressions. Compared with RefCOCO, the words indicate the absolute position, like 'left' and 'right', are not included in the language expressions from this dataset. It is also officially split into a train set with 120,191 expressions, a validation set with 10,758 expressions, a testA set with 5,726 expressions and a testB set with 4,889 expressions. 

\textit{RefCOCOg}~\cite{mao2016generation} has 25,799 images with 49,856 referred objects and expressions. In this dataset, each image contains 2 to 4 instances for the referred categories, and each instance are with an area more than 5\% of the whole image. There are two commonly used split protocols for this dataset. One is RefCOCOg-google~\cite{mao2016generation}, and the other is RefCOCOg-umd~\cite{nagaraja2016modeling}. We report our performance on both RefCOCOg-google (val-g) and RefCOCOg-umd (val-u and test-u) to make comprehensive comparisons.

\textit{ReferItGame}~\cite{kazemzadeh2014referitgame} includes 20,000 images collected from the SAIAPR-12 dataset~\cite{escalante2010segmented}, and each image has one or a few regions with  corresponding referring expressions. This dataset is divided into three subsets, \emph{i.e.}, a train set with 54,127 referring expressions, a validation set with 5,842 referring expressions and a test set with 60,103 referring expressions. We use the validation set for ablation studies and compare our method with others on the test set.

\textit{Flickr30K Entities}~\cite{plummer2017flickr30k} is built on the original Flickr30K \cite{young2014image}, through introducing short region phrase correspondence annotations. It contains 31,783 images with 427K referred entities. We follow the previous works~\cite{plummer2017flickr30k,plummerCITE2018,wang2019learning,yang2020improving} to separate the these images into 29,783 for training, 1000 for validation, and 1000 for testing.

\subsection{Experimental Setup} \label{sec4:impl}

\noindent{\bf Inputs.} 
The input image size is set as $640\times640$. When performing image resizing, we keep the original aspect ratio of each image. The longer edge of an image is resized to 640, while the shorter one is padded to 640 with the mean value of RGB channels. The maximum length of language expressions is set as 40 for RefCOCOg, and set as 20 for other datasets. We cut off the language query if its length is longer than maximum length (leaving one position for the \texttt{[CLS]} token and one position for the \texttt{[SEP]} token). Otherwise, we pad empty tokens after \texttt{[SEP]} token to make the input length consistent in each batch. The padded tokens are recorded with a mask, and will not influence the output.

\noindent{\bf Network Architecture.} In our preliminary framework, we use ResNet-50 or ResNet-101~\cite{he2016deep} as the convolutional backbone network. The output of ResNet is 32 times downsampling against the input image. Transformer encoder layers in the preliminary framework follow the post-normalization structure. The embedding dimension is set as 256, the head number of multi-head attention modules is set as 8, and the hidden dimension in FFN is set as 2048.

In our advanced framework, the patch embedding layer divides a $640\times640$ image into patches of equal size, each with $16\times16$ pixels. These patches are flattened into vectors and fed into a fully connected layer to generate the initial state of visual tokens. To encode position information into visual tokens, learnable position embedding is added to each visual token. Pariticularly, the resolution of position embedding in the standard ViT model for image classification is $14\times14$, as the input size is only $224\times224$. To make the position embedding matches the input size of our method, we follow~\cite{chen2021simple} to resize it to $40\times40$ with bicubic interpolation. We experiment with three representative models of ViT series, \emph{i.e.}, ViT-tiny, ViT-small and ViT-base, which is for vision feature encoding. The configuration of LViT varies according to the ViT variety. The feature dimension of our language adapter and language prompter changes according to the ViT model. Specifically, for ViT-tiny, the embedding dimension is 192, the head number of multi-head attention module is 3, the hidden dimension in FFN is 768. For ViT-small, the embedding dimension is 384, the head number of multi-head attention module is 6, the hidden dimension in FFN is 1536. For ViT-small, the embedding dimension is 768, the head number of multi-head attention module is 12, the hidden dimension in FFN is 3072. No matter what ViT model is leveraged, the hidden dimension in regression MLP is kept unchanged (\emph{i.e.}, 256 as that in preliminary version) to eliminate the influence of feature dimension in prediction head.

\noindent{\textbf{Training and Inference Details.}} 
Our model is end-to-end optimized with AdamW optimizer, with batch size set as 32. We apply several widely adopted data augmentation techniques~\cite{liao2020real,yang2020improving,yang2019fast,deng2021transvg,li2021referring} at the training stage. In TransVG, we set the initial learning rate of fusion module and prediction head to $10^{-4}$ and that of the vision branch and language branch to $10^{-5}$. The weight decay is $10^{-4}$. On RefCOCO, RefCOCOg and ReferItGame datasets, TransVG is trained for 90 epochs with a learning rate dropped by a factor of 10 after 60 epochs. On Flickr30K Entities, TransVG is trained for 60 epochs, with a learning rate dropping after 40 epochs. On RefCOCO+, TransVG is trained for 180 epochs, with a learning rate drops after 120 epochs. For TransVG++, we set the initial learning rate of language branch and the parameters initialized from the pre-trained ViT model to $10^{-5}$, and that of other parameters to $10^{-4}$. Thanks to removing the stand-alone fusion Transformer, the total training epochs of TransVG++ is reduced to 60 epochs. The learning rate drops by 10 times since the 45-th epoch. We follow the common practice~\cite{yang2019fast,yang2020improving,liu2020learning,deng2021transvg,li2021referring} to initialize the parameters of language branch with a pre-trained BERT$_\text{BASE}$ model~\cite{devlin2018bert}, and to initialize the parameters of vision branch with models trained on MSCOCO~\cite{lin2014microsoft} (the overlapping images between the training set of MSCOCO and validation/test sets of RefCOCO/RefCOCO+/RefCOCOg datasets are removed when performing pre-training). We use DETR model~\cite{carion2020end} for vision branch initialization in TransVG, and instead use Mask R-CNN~\cite{he2017mask} in TransVG++, since these two initialization options achieve comparable performance, while the training epochs of Mask R-CNN is remarkably less than DETR. 

Since our proposed frameworks directly output the box coordinates, there is no post-processing during inference.

\subsection{Ablative Experiments}\label{sec4:analysis}

\subsubsection{Ablation Studies on TransVG}

In this section, we conduct ablative experiments to investigate the effectiveness of each component in our TransVG framework. We exploit ResNet-50 as the backbone network of the vision branch. All of the compared models are trained for 90 epochs on ReferItGame~\cite{kazemzadeh2014referitgame} for fair comparison.

    \begin{table}[t]
    
    \centering
    \footnotesize
    \caption{Ablative experiments on the design of \texttt{[REG]} token's initial state in TransVG. The evaluation metric is the accuracy (\%) on ReferItGame validation and test set. We compare our default strategy to exploit learnable embedding as the initial state of \texttt{[REG]} token with other options that generate the initial state from vision/language tokens out of the corresponding branch.}
    \renewcommand\arraystretch{1.0}
    \begin{center}
		\scalebox{0.95}[0.95]{
    		\setlength\tabcolsep{12pt}
    		\begin{tabular}{lcc}
    			\toprule[1.2pt]
    			Initial State of \texttt{[REG]} Token & Acc@val & Acc@test \\
    			\midrule[0.8pt]
    			Average pooled vision tokens & 71.37 & 69.27 \\
    			Max pooled vision tokens & 70.91 & 69.11 \\
    			Average pooled language tokens & 69.96 & 68.15 \\
    			Max pooled language tokens & 70.37 & 68.46 \\
    			Sharing with \texttt{[CLS]} token & 70.84 & 69.01 \\
    			\rowcolor{Gray1}
    			$\text{Learnable embedding}^*$ & $\bm{72.50}$ & $\bm{69.76}$\\
    			\bottomrule[1.2pt]
    		\end{tabular}
		}
    \end{center}
    \label{tab:ablative_token}
    \end{table}

    \begin{table}[t]
    \centering
    \footnotesize
    \caption{Ablative experiments on the vision branch and language branch in our TransVG. The performance is evaluated on the test set of ReferItGame in terms of accuracy (\%). ``Tr.'' represents Transformer.}
    \renewcommand\arraystretch{1.0}
    \begin{center}
    	\scalebox{0.95}[0.95]{
    		\setlength
    		\tabcolsep{10pt}
    		\begin{tabular}{cc|cc|c}
    			\toprule[1.2pt]
    			\multicolumn{2}{c|}{Vision Branch} & \multicolumn{2}{c|}{Language Branch} & \multirow{2}{*}{Accuracy (\%)} \\
    			w/o Tr.  & w/ Tr. & w/o Tr. & w/ Tr. & \\
    			\midrule[0.8pt]
    			\checkmark & & \checkmark &  & 64.24\\
     			\checkmark & &  &\checkmark  & 66.78~(+3.54) \\
    			 & \checkmark & \checkmark &  & 68.48~(+4.24) \\
    			 \rowcolor{Gray1}
    			& \checkmark & & \checkmark & $\bm{69.76}$~(+5.52) \\
    			\bottomrule[1.2pt]
    		\end{tabular}
    	}
    \end{center}
    \label{tab:ablative_tr}
    \vspace{-0.25cm}
    \end{table}

\noindent{\textbf{Design of \texttt{[REG]} Token's Initial State.}}
In Table~\ref{tab:ablative_token}, we report the abation study on how to obtain the initial state of \texttt{[REG]} token. Specifically, we compare our learnable embedding with five other options to generate the initial state of \texttt{[REG]} token (\emph{i.e.}, the embedding $g_r$ appended to vision embedding and language embedding as in Equation~\eqref{eq:token}). We detail these designs and analysis as follows:
\begin{itemize}[nolistsep]
    \item[---] \textit{Average pooled vision tokens.} We perform average pooling over the tokens out of vision branch, and exploit the average-pooled embedding as the initial state.
    \item[---] \textit{Max pooled vision tokens.} We perform max pooling over the tokens out of vision branch, and exploit the max-pooled embedding as the initial state.
    \item[---] \textit{Average pooled language tokens.} Similar to the first option, but using the language tokens.
    \item[---] \textit{Average pooled language tokens.} Similar to the second choice, but using the language tokens.
    \item[---] \textit{Sharing with \texttt{[CLS]} token.} We use the \texttt{[CLS]} token of language embedding to play the role of \texttt{[REG]} token. In this setting, the \texttt{[CLS]} token out of the V-L module is fed into the prediction head.
    \item[---] \textit{Learnable embedding*.} This is our default setting by randomly initializing the \texttt{[REG]} token embedding at the beginning of the training stage. And the parameters of this embedding are optimized with the whole model.
\end{itemize}

Our proposed design to exploit a learnable embedding achieves 72.50\% and 69.76\% on the validation test set of ReferItGame, which is the best performance among all the designs. Typically, the initial \texttt{[REG]} token of other designs is either generated from vision or language tokens, which involves biases to the specific prior context of the corresponding modality. In contrast, the learnable embedding tends to be more equitable and flexible when performing relation reasoning in the vision-language Transformer.

\noindent{\textbf{Design of Transformers in Each Branch.}}
We study the role of the Transformers in the vision branch and the language branch, and report the comparison in Table~\ref{tab:ablative_tr}. The baseline model without the uni-modal Transformer in both branch reports an accuracy of 64.24\%. When we only attach Transformer to each vision branch or language branch, an improvement of 68.48\% and 66.78\% are achieved, respectively. With the complete architecture, the performance is further boosted to 69.76\% on the ReferIt test set. This result demonstrates the essential of Transformers in the vision branch and language branch to capture intra-modality global context before performing multi-modal fusion.

\begin{table}[t]
    \centering
    \footnotesize
    \caption{Performance comparison between different strategies to perform vision-languge fusion. Two representative datasets, \emph{i.e.}, ReferItGame and RefCOCOg-google, are exploited for evaluation. We report the accuracy (\%) on validation sets of these two datasets.}
    \renewcommand\arraystretch{1.0}
	
    \begin{center}
		\scalebox{0.95}[0.95]{
    		\setlength\tabcolsep{15pt}
    		\begin{tabular}{lcc}
    			\toprule[1.2pt]
    			Fusion Strategies & ReferIt & RefCOCOg-g \\
                \midrule[0.8pt]
                \textbf{\textit{Preliminary:}} &  &  \\
                \qquad 6 encoder layers & 71.00 & 67.24\\
    			\midrule[0.8pt]
    			\textbf{\textit{Language Prompter:}} \\
    			\qquad MHSA  & 71.39 & 68.46 \\
    			\qquad Encoder layer & 71.38 & 69.07 \\
    			\qquad Linear  & 71.62  & 69.09  \\
    			\qquad FFN  & 71.42 & 69.44 \\
    			\midrule[0.8pt]
    			\textbf{\textit{Language Adapter:}} \\
    			\qquad Average pooling & 72.01 & 69.49  \\ 
    			\qquad Max pooling & 72.26 & 68.96 \\ 
    			\qquad $\texttt{[CLS]}$ only & 72.79 & 69.54 \\
    			\rowcolor{Gray1}
    			\qquad MHCA$^*$ & $\bm{73.17}$ & $\bm{70.60}$\\
    			
    			\bottomrule[1.2pt]
    		\end{tabular}
		}
    \end{center}
    \label{tab:ablative_fusion}
     \vspace{-0.15cm}
    \end{table}

\subsubsection{Ablation Studies on TransVG++}

In this section, we conduct ablation studies on TransVG++. Since the language branch remains unchanged, these studies focus on analysing LViT. All of the compared models exploit ViT-tiny as the backbone network of LViT, and are trained for 60 epochs for fair comparison.

\noindent{\bf Degisn of Fusion Strategies.} In Table~\ref{tab:ablative_fusion}, We present ablation studies on the effectiveness of different vision-language fusion strategies. These models are compared on the validation set of representative ReferItGame and RefCOCOg-google datasets. Our preliminary fusion strategy is to concatenate vision tokens and language tokens together, and fed them into 6 stand-alone Transformer encoder layers. By leveraging this preliminary strategies, the model achieves 71.00\% accuracy and 67.24\% accuracy on ReferItGame and RefCOCOg-google, respectively. 

Generally speaking, models capitalizing on language prompters and language adapters, no matter what internal structure is used, achieve better accuracy than the stand-alone Transformer encoder layers, which demonstrates the advantage of fusion in backbone against fusion with external modules.
We experiment with four different types of language prompters to generate language prompt tokens, including a simple linear layer, a feed forward network (FFN), a Transformer encoder layer, and a multi-head self-attention (MHSA) module. Language prompters with these four structures work comparably on ReferItGame dataset, and the model with an FFN works the best on RefCOCOg-google, achieving 69.44\% accuracy. In experiments with language adapters, we evaluate four alternative designs to obtain language adapt tokens, \emph{i.e.}, average pooling over language tokens, max pooling over language tokens, taking linear transformation over \texttt{[CLS]} token, and our default setting with a multi-head cross-attention (MHCA) module. The former three designs can be regarded as the degeneration of language adapters with MHCA. Specifically, \textit{avg pooling} sets all the attention weights equally, \textit{max pooling} select one token for each channel to set its weight as 1 and that of others as 0, and \textit{[CLS] only} sets attention weights of \texttt{[CLS]} token as 1 and that of others as 0. These three designs achieve similar performance with using MHCA module when processing language expressions with simple phrases like that in ReferItGame, while lags behind using MHCA module when it comes to complicated sentences as collected in RefCOCOg-google.

\begin{table}[t]
    \centering
    \footnotesize
    \caption{Ablative experiments on the location of language conditioned visual encoder layers. The performance is evaluated on RefCOCOg-google dataset in terms of accuracy (\%). We use ViT-tiny backbone by default.}
    \renewcommand\arraystretch{1.0}
    \begin{center}
    		\setlength\tabcolsep{28pt}
    		\begin{tabular}{lc}
    			\toprule[1.2pt]
    			Location & Accuracy (\%) \\
    			\midrule[0.8pt]
    			First 4 layers & 68.26\\
    			Last 4 layers & 70.53\\ 
    			\rowcolor{Gray1}
    			Evenly 4 layers$^*$  & $\bm{70.60}$ \\
    			\bottomrule[1.2pt]
    		\end{tabular}

    \end{center}
    \label{tab:ablative_location}
    \end{table}

\begin{table}
		\centering
		\footnotesize
		\caption{Ablative experiments on the number of language conditioned vision encoder layers. We use ViT-tiny backbone by default, and evaluate these models on RefCOCOg-google.}
		\renewcommand\arraystretch{1.0}
		\begin{center}
				\setlength\tabcolsep{36pt}
				\begin{tabular}{cc}
					\toprule[1.2pt]
					Number  & Accuracy (\%) \\
					\midrule[0.8pt]
					2 &  70.13 \\ 
					3 & 70.34 \\
					\rowcolor{Gray1}
					4$^*$ & $\bm{70.60}$ \\
					6 & 70.58 \\
					12 & 70.30 \\
					\bottomrule[1.2pt]
				\end{tabular}
	
		\end{center}
		\label{tab:ablative_number}
		\end{table}
		
As empirically shown in the table, the best choice is to use language adapters with MHCA for language adapt token generation, which achieves 73.17\% on ReferItGame and 70.60\% on RefCOCOg-google, outperforming the preliminary strategy by 2.17\% and 3.36\%, respectively. We leverage this setting as our default one, and report the performance of models with this setting in our main results.

\begin{table}[t]
	\caption{Ablative experiments on improvements from TransVG to TransVG++.  Both accuracy on validation set of RefCOCOg-google and model size are compared.}
	\footnotesize
	\begin{center}
	    \renewcommand\arraystretch{1.2}
		\scalebox{0.9}[0.9]{
			\setlength
			\tabcolsep{3.5pt}
			\begin{tabular}{c | c | c | cc | c }
				\hline
				\toprule[1.2pt]
				& \multirow{2}{*}{Vision Feature} & \multirow{2}{*}{V-L Fusion} & \multicolumn{2}{c|}{Model Size (M)} & Acc.\\ 
				& &  & Vision & Fusion & (\%) \\
				\midrule[0.8pt]
				(a) &  R-50 + 6 enc. layers  & 6 enc. layers &30.18 & 7.52 & 66.35 \\
				(b) &  ViT-tiny  & 6 enc. layers & 5.53 & 7.52 & 67.24 \\
				(c) & R-50-DC5+ 6 enc. layers & 6 enc. layers &30.18 & 7.52 & 65.96 \\
				(d) &  R-50 + 6 enc. layers & language adapter & 31.87 & 1.50 & 68.72\\
				\rowcolor{Gray1}
				(e) &  ViT-tiny  & language adapter & 5.53 & 0.57 & $\bm{70.60}$ \\
				\bottomrule[1.2pt]
			\end{tabular}
		} 
	\end{center}
	\label{tab:lvit_ablation}
\end{table}

\begin{table*}[t]
	\caption{Performance comparisons of our preliminary version (TransVG) and advanced version (TransVG++) frameworks with other state-of-the-art methods on RefCOCO~\cite{yu2016modeling}, RefCOCO+~\cite{yu2016modeling} and RefCOCOg~\cite{mao2016generation} in terms of accuracy (\%). For comprehensive comparison, we evaluate the performance of TransVG with ResNet-50 and ResNet-101 as the backbone network, and evaluate the performance of TransVG++ with ViT-tiny, ViT-small and ViT-base as the backbone network. We add "$\dag$" to Refformer~\cite{li2021referring} to indicate this method needs more annotations for training. 
	}
	
	\footnotesize

	\begin{center}
	    \renewcommand\arraystretch{1.0}
		\scalebox{0.95}[0.95]{
			\setlength
			\tabcolsep{10pt}
			\begin{tabular}{l | c | c c c | c c c | c c c }
				\hline
				\toprule[1.2pt]
				\multirow{2}{*}{Models} & \multirow{2}{*}{Backbone} & \multicolumn{3}{c|}{RefCOCO} & \multicolumn{3}{c|}{RefCOCO+} & \multicolumn{3}{c}{RefCOCOg} \\ 
				
				&  & val & testA & testB & val & testA & testB & val-g & val-u & test-u \\
				\midrule[0.8pt]
				
				\textbf{\textit{Two-stage:}} & & & & & & & & & &\\
				CMN~\cite{hu2017modeling}  & VGG16 & - & 71.03 & 65.77 & - & 54.32 & 47.76 & 57.47 & - & - \\
				VC~\cite{zhang2018grounding} & VGG16 & - & 73.33 & 67.44 & - & 58.40 & 53.18 &62.30 & - & - \\
				ParalAttn~\cite{zhuang2018parallel} & VGG16 & - & 75.31 & 65.52 & - & 61.34 & 50.86 & 58.03 & - & - \\
				LGRANs~\cite{wang2019neighbourhood}  & VGG16 & - & 76.60 & 66.40 & - & 64.00 & 53.40 & 61.78 & - & - \\
				MAttNet~\cite{yu2018mattnet}  &	ResNet-101 & 76.65 & 81.14 & 69.99 & 65.33 & 71.62 & 56.02 & -  & 66.58 & 67.27 \\
				DGA~\cite{yang2019dynamic}  & ResNet-101 & - & 78.42 & 65.53 & - & 69.07 & 51.99 & - & - & 63.28 \\ 
				RvG-Tree~\cite{hong2019learning}  & ResNet-101 & 75.06 & 78.61 & 69.85 & 63.51 & 67.45 & 56.66 & - & 66.95 & 66.51 \\
				CMRE~\cite{yang2020relationship} & ResNet-101 & - & 82.53 & 68.58 & - & 75.76 & 57.27 & - & - & 67.38 \\
				NMTree~\cite{liu2019learning}  & ResNet-101 & 76.41 & 81.21 & 70.09 & 66.46 & 72.02 & 57.52 & 64.62 & 65.87 & 66.44 \\
				CM-A-E~\cite{liu2019improving} & ResNet-101 & 78.35 & 83.14 & 71.32 & 68.09 & 73.65 &	58.03 & - & 67.99 & 68.67  \\

				\midrule[0.8pt]

				\textbf{\textit{One-stage:}} &  & & & & & & & & &\\
				SSG~\cite{chen2018real}  &  DarkNet-53 & - & 76.51 & 67.50 & - & 62.14 & 49.27 & 47.47 & 58.80 & - \\  
				FAOA~\cite{yang2019fast} & DarkNet-53 & 72.54 & 74.35 & 68.50 & 56.81 & 60.23 & 49.60 & 56.12 & 61.33 & 60.36 \\
				RCCF~\cite{liao2020real}  & DLA-34 & - & 81.06 & 71.85 & - & 70.35 & 56.32 & -  & - & 65.73 \\
				ReSC-Large~\cite{yang2020improving} & DarkNet-53 & 77.63 & 80.45 & 72.30 & 63.59 & 68.36 & 56.81 & 63.12 & 67.30 & 67.20 \\
				
				\midrule[0.8pt]
				
				\textbf{\textit{Transformer-based:}} &  & & & & & & & & &\\
				Refformer$^\dag$~\cite{li2021referring} & ResNet101 & 82.23 & 85.59 & 76.57 & 71.58 & 75.96 & 62.16 & - & 69.41 & 69.40 \\
				VGTR~\cite{du2022vgtr} & ResNet101 & 79.30 & 82.16 & 74.38 & 64.40 & 70.85 & 55.84 & 64.05 & 66.83 & 67.28 \\
				
				\midrule[0.8pt]
				
				\textbf{\textit{Ours:}} &  & & & & & & & & &\\
				
				TransVG & ResNet-50 & 80.49 & 83.28 & 75.24 & 66.39 & 70.55 & 57.66 & 66.35 & 67.93 & 67.44 \\
				
				TransVG & ResNet-101 & 80.83 & 83.38 & 76.94 & 68.00 & 72.46 & 59.24 & 68.03 & 68.71 & 67.98 \\
				
				TransVG++ & ViT-tiny  & 82.93 & 85.45 & 77.67 & 69.17 & 74.46 & 59.59 & 70.60 & 70.98 & 71.83 \\
				
				TransVG++ & ViT-small & 85.24 & 87.50 & 80.46 & 73.73 & 79.21 & 63.56 & 73.43 & 74.78 & 74.77 \\ 
				
				TransVG++ & ViT-base & $\bm{86.28}$ & $\bm{88.37}$ & $\bm{80.97}$ & $\bm{75.39}$ & $\bm{80.45}$ & $\bm{66.28}$ & $\bm{73.86}$ & $\bm{76.18}$ & $\bm{76.30}$ \\ 
				
				\bottomrule[1.2pt]
			\end{tabular}
		} 

	\end{center}
	\label{tab:refcoco_results}
\end{table*}

\noindent{\bf Location of Language Conditioned Vision Encoder Layers.} In Table~\ref{tab:ablative_location}, we study where language conditioned vision encoder layers should be located (\emph{i.e.}, applying language adapters to which vision encoder layers). The results are evaluated on RefCOCOg-google dataset. Our default configuration is to evenly place the language adapters at the 3rd, 6th, 9th and 12th vision encoder layers. We compare two different configurations to place the language adapters at the first 4 encoder layers or the last 4 encoder layers, which attempts to perform multi-modal fusion at the early stage or at the late stage, respectively. As shown in the table, our default setting ranks the first among these configurations.

\noindent{\bf Number of Language Conditioned Vision Encoder Layers.} Table~\ref{tab:ablative_number} presents our ablation studies on how many language conditioned vision encoder layers should be exploited. When increasing the number of language conditioned vision encoder layers from 2 to 4, the accuracy is consistently improved from 70.13\% to 70.60\%. The performance is not shown to benefit from more language conditioned vision encoder layers once the number achieves 4. Therefore, we empirically set the default number as 4.

\subsubsection{Improvements from TransVG to TransVG++}

In this section, we verify the improvements from TransVG to TransVG++, \emph{i.e.}, upgrading the vision branch to a fully Transformer-based architecture and removing the stand-alone V-L fusion module. 
Table~\ref{tab:lvit_ablation} details five varieties of architecture configures, together with the model size and accuracy. Since the language branch of all the models follow the same configuration, it is omitted in this table. For clearer comparison, we split LViT into ViT and language adapters in this table. These models are evaluated on the validation set of RefCOCOg-google~\cite{mao2016generation} dataset.

\textit{Model (a)} is the preliminary baseline, \emph{i.e.}, TransVG. It exploits ResNet-50 and 6 following Transformer encoder layers for vision feature extraction and 6 stand-alone Transformer encoder layers for multi-modal fusion. In this model, the vision branch has 30.18M parameters and the vision-language fusion module has 7.52M parameters. It achieves 66.35\% accuracy on the validation set of RefCOCOg-google.

\textit{Model (b)} replaces ResNet-50 and following Transformer encoder layers in model (a) with ViT-tiny, reducing the parameters of vision feature extraction model to 5.53M. The model size of (b) is more than 5 times smaller than that of (a), while the accuracy of (a) achieves 67.24\%, outperforming (a) by 0.89\%. Note that ViT-tiny's capability for image classification and object detection lags behind ResNet-50, which shows the inconsistency between vision perception and vision-language understanding. Besides, the better performance of (b) demonstrates the advantage of fully Transformer-based framework.

\textit{Model (c)} leverages dilation convolution~\cite{chen2014semantic} in the last stage of ResNet-50 backbone. By setting the dilation ratio as 2, the downsampling rate of ResNet-50 reduces from 32 to 16. Therefore, the input vision tokens to the following Transformer encoder layers are with the same number as that of patches for a ViT backbone (\emph{i.e.}, 1600). As observed, the performances drops to 65.96\%, which demonstrates that the performance difference between model (a) and model (b) is not caused by token numbers, but due to the powerful ViT backbone and fully Transformer-based architecture.

\textit{Model (d)} uses the same structure as model (a) for vision feature extraction, but capitalizes on language adapters for multi-modal fusion. In this model, a language adapter is added to each Transformer encoder layer, converting the original 6 vision encoder layers for vision encoding to language conditioned vision encoder layers. To match the feature dimension of Transformers in model (a), in the language adapter, the embedding dimension is set to 256, and head number of cross-attention module is set to 8. The performance of model (d) reaches 68.72\%, outperforming model (a) by 2.37\% and involving significant less parameters. This result demonstrates the significance of removing stand-alone vision-language fusion blocks and instead enabling fusion in the vision encoder, which makes the model easier for optimization.

\textit{Model (e)} is the default configuration of TransVG++ with language adapter strategy. By removing the stand-alone V-L fusion module composed of 6 Transformer encoder layers and capitalizing on language adapters to fuse language expression information into the vision tokens at intermediate layers of a ViT backbone network, (e) only introduces 0.57M extra parameters. Besides, the computation costs of performing multi-modal fusion is reduced from 7.38G FLOPs in (b) to 0.92G FLOPs in (e). By making the best of fully Transformer-based structure and the effective fusion-in-vision-encoder strategy, model (e) boosts the performance of baseline (a) from 66.35\% to 70.60\% accuracy.

\subsection{Main Results and Comparisons}\label{sec4:overall_comp}

\subsubsection{RefCOCO/RefCOCO+/RefCOCOg}
{\bf Comparison with Two-stage and One-stage Methods.}
Table~\ref{tab:refcoco_results} reports the performance of our proposed TransVG (preliminary version) and TransVG++ (advanced version), together with other competitive two-stage and one-stage methods on the widely adopted RefCOCO, RefCOCO+ and RefCOCOg datasets. To make comprehensive comparison, we present the results of TransVG with ResNet-50 and ResNet-101 backbone network, and that of TransVG++ with ViT-tiny, ViT-small and ViT-base. Here, our proposed TransVG++ adopts language adapter strategy with the default MHCA setting to integrate language information into vision tokens out of ViT's intermediate layers. 
With ResNet-101 backbone, our TransVG consistently outperforms all the one-stage methods on all the subsets and splits. Besides, TransVG outperforms the strongest two-stage competitor (\emph{i.e.}, CM-A-E~\cite{liu2019improving}) by a remarkable margin of 5.62\% accuracy on the testB set of RefCOCO, and achieves comparable performance on other datasets. 

In particular, we find the recurrent architecture in ReSC shares the same spirit with our stacking architecture in the visual-linguistic Transformer that fuses the multi-modal context in multiple rounds. However, in ReSC, recurrent learning is only performed to construct the language sub-query, and this procedure is isolated from the sub-query attended visual feature modulation. In contrast, our TransVG embeds the visual and linguistic embedding into a common semantic space by homogeneously performing intra- and inter-modality context reasoning. The superiority of our performance empirically demonstrates that the complicated multi-modality fusion module can be replaced by a simple stack of Transformer encoder layers.

Among all the methods, our advanced TransVG++ series achieve the best grounding accuracy on all the benchmarks. Even with ViT-tiny backbone, which has only about 5.53M parameters, TransVG++ outperforms most of the competitors by a remarkable margin. Such a result validates the effectiveness of our proposed strategy to preserve the power of ViT and fuse the language information into vision features by injecting the language expression tokens into the intermediate vision Transformer encoder layers. When upgrading ViT-tiny to models with larger capacity, \emph{i.e.}, ViT-small and ViT-base, we observe consistent performance improvements, which shows the unified fully Transformer design facilitates TransVG++ to benefit from the future advances in vision Transformers. Besides, the training process of TransVG++ convergences much faster than TransVG. In TransVG, the core vision-language Transformer devised for multi-modal fusion is trained from scratch, making it hard to be optimized on the limited visual grounding data. On the contrary, TransVG++ removes the stand-alone fusion Transformer, avoiding this problem. Specifically, to achieve the reported performance on the challenging RefCOCO+ dataset, the preliminary version model, \emph{i.e.}, TransVG, is trained for 180 epochs, while we only optimize TransVG++ for 60 epochs on the same dataset.

\begin{table}
	\caption{Comparisons with state-of-the-art methods on the test set of  ReferItGame~\cite{kazemzadeh2014referitgame} and Flickr30K Entities~\cite{plummer2017flickr30k} datasets in terms of accuracy (\%). }
	\footnotesize
	\begin{center}
	    \renewcommand\arraystretch{1.0}
		\scalebox{0.95}{
			\setlength
			\tabcolsep{8pt}
			\begin{tabular}{l | c | c | c  }
				\toprule[1.2pt]
				Models & Backbone & ReferItGame & Flickr30K \\
				
				\midrule[0.8pt]
				\textbf{\textit{Two-stage:}}  &  &  &\\
				CMN~\cite{hu2017modeling} & VGG16 & 28.33 & -\\
				VC~\cite{zhang2018grounding} & VGG16 & 31.13 & - \\
				MAttNet~\cite{yu2018mattnet} &	ResNet-101  & 29.04 & - \\
				Similarity Net~\cite{wang2019learning} & ResNet-101 & 34.54 & 60.89\\
				CITE~\cite{plummerCITE2018}  & ResNet-101  & 35.07 & 61.33\\
				PIRC~\cite{kovvuri2018pirc} & ResNet-101 & 59.13 & 72.83 \\
				DDPN~\cite{yu2018rethinking} & ResNet-101 &  63.00 & 73.30\\
				LCMCG~\cite{liu2020learning} & ResNet-101 & - & 76.74 \\
				\midrule[0.8pt]
				\textbf{\textit{One-stage:}}  &  & &\\
				SSG~\cite{chen2018real} & DarkNet-53 & 54.24 & - \\
				ZSGNet~\cite{sadhu2019zero} & ResNet-50 & 58.63 & 63.39\\
				FAOA~\cite{yang2019fast} & DarkNet-53 & 60.67 & 68.71\\
				RCCF~\cite{liao2020real} & DLA-34 & 63.79 & -\\
				ReSC-Large~\cite{yang2020improving} & DarkNet-53 & 64.60 & 69.28\\
				
				\midrule[0.8pt]
				\textbf{\textit{Ours:}}  &  &   &\\
				TransVG & ResNet-50 &  69.76 & 78.47 \\
				TransVG & ResNet-101 &  70.73 & 79.10 \\
				
				TransVG++ & ViT-tiny & 70.85 &  78.19 \\
				TransVG++ & ViT-small & 73.55 & 80.36\\
				TransVG++ & ViT-base & $\bm{74.70}$ & $\bm{81.49}$\\
				\bottomrule[1.2pt]
			\end{tabular}
		} 
	\end{center}
	\label{tab:referit_results}
\end{table}

\noindent{\bf Comparison Among Transformer-based Methods.} In Table~\ref{tab:refcoco_results}, we also present a comparison between our framework with other Transformer-based methods, \emph{i.e.}, Refformer~\cite{li2021referring} and VGTR~\cite{du2022vgtr}. Compared with our TransVG, the following work Refformer adds an query encoder and a visual context decoder to the multi-modal fusion module, and jointly optimizes the model with both annotations of referring expression comprehension (visual grounding) and that of referring expression segmentation. Thanks to involving extra annotations from referring expression segmentation at the training stage, the problem of training the core core multi-modal fusion module on limited data can be partially alleviated. Alternatively, our TransVG++ waives the requirements of stand-alone fusion module, not only introducing less parameters, but also addressing the aforementioned problem in a simpler and more effective way. As shown in the table, our TransVG++ outperforms Refformer on all the subsets of the evaluated datasets.

\subsubsection{ReferItGame}
To further validate the merits of our proposed framework, we also conduct experiments on ReferItGame dataset, and report the performance on the test set. As shown in Table~\ref{tab:referit_results}, both the preliminary TransVG and the advanced TransVG++ largely outperform previous two-stage and one-stage methods.  Specifically, with ResNet-50 backbone, TransVG achieves 69.76\% top-1 accuracy and outperforms ZSGNet~\cite{sadhu2019zero} with the same backbone network by 11.13\%. By replacing ResNet-50 with a stronger ResNet-101, the performance boosts to 70.73\%, which is 6.13\% higher than the strongest competitor ReSC-Large for one-stage methods and 7.73\% higher than the strongest competitor DDPN for two-stage methods, respectively. As the first fully-Transformer model, TransVG++ capitalizing on ViT-base further outperforms TransVG with ResNet-101 by 3.97\%, showing the effectiveness of our proposed framework to integrate language information into the vision encoders.

Among the competitors, MAttNet~\cite{yu2018mattnet} is the most representative method that devises multi-modal fusion modules with re-defined structures (\emph{i.e}, modular attention networks to separately model subject, location and relationship). When we compare our model with MAttNet in Table~\ref{tab:referit_results} and Table~\ref{tab:refcoco_results}, we can find that MAttNet shows comparable results to our preliminary version on RefCOCO/RefCOCO+/RefCOCOg, but lags behind ours on RefeItGame. The reason is that the pre-defined relationship in multi-modal fusion modules makes it easy to overfit to specific datasets (\emph{e.g.}, with  specific  scenarios,  query  lengths,  and  relationships). Our proposed framework effectively avoids this problem by establishing intra-modality and inter-modality correspondence with the flexible and adaptive attention mechanism.

\subsubsection{Flickr30K Entities}
Table~\ref{tab:referit_results} also reports the performance of our framework on the Flickr30K Entities test set. On this dataset, our TransVG achieves 79.10\% top-1 accuracy with a ResNet-101 backbone network. Our TransVG++ with a ViT-base backbone network further boosts the accuracy to 81.49\%, surpassing the recently proposed Similarity Net~\cite{wang2019learning}, CITE~\cite{plummerCITE2018}, DDPN~\cite{yu2018rethinking}, ZSGNet~\cite{sadhu2019zero}, FAOA~\cite{yang2019fast}, and ReSC-Large~\cite{yang2020improving} by a remarkable margin (\emph{i.e.}, 4.75\% absolute improvement over the previous state-of-the-art record).

\subsection{Discussion on Model Size and Computation Costs} \label{sec4:discussion}

In this section, we perform analysis on model size and computation costs of our TransVG++ framework. As shown in Table~\ref{tab:lvit_model_size}, the parameters in language branch make up the major part of TransVG++, \emph{i.e.}, 94.4\% in TransVG++ (tiny), 81.6\% in TransVG++ (small) and 53.3\% in TransVG++ (base). Even with the largest ViT-base backbone, the number of parameters in vision branch (\emph{i.e.}, LViT) is less than that in the language branch. Note that we use language Transformer with the same configuration in language branch and regression MLP with the same hidden dimension in prediction head, no matter what vision backbone is leveraged. 
The trivial differences in model size of these two components are due to the different embedding dimension for matching different ViT models. 
Compared with the size of the whole model, the number of parameters introduced by our proposed language adapter is negligible.
Specifically, in TransVG++ (tiny), the language adapter only introduces 0.57M parameters, which only accounts for 0.49\% of the total number of parameters. This comparison demonstrates the parameter efficiency of our proposed method.

\begin{figure}[t]
	\centering
	\includegraphics[width=1.0\linewidth]{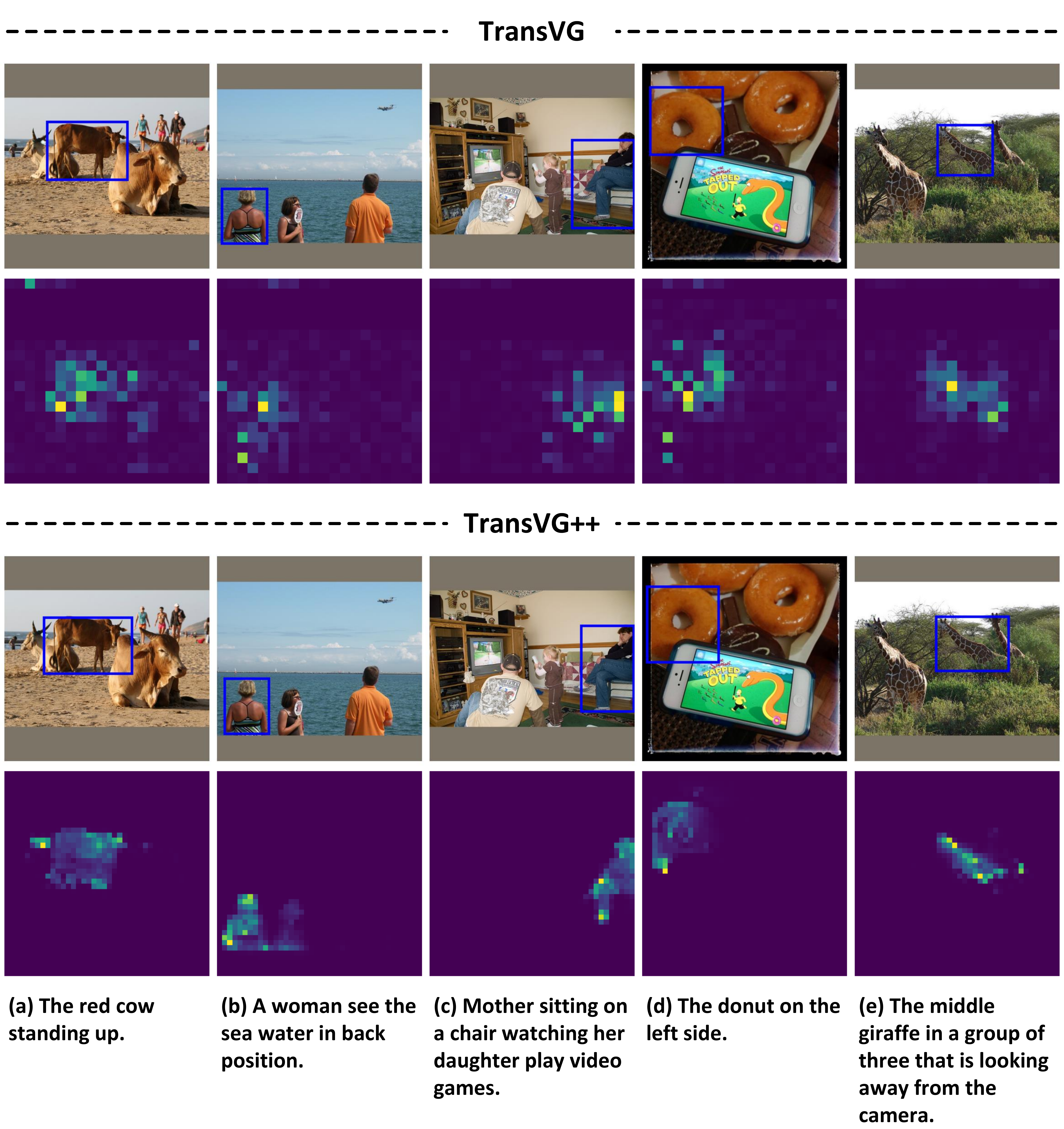}
	\caption{Qualitative results of our proposed TransVG and TransVG++ frameworks on the RefCOCOg test set (better viewed in color). We show both the predicted bounding box and the [REG] token's attention over vision tokens from the last fusion block of each framework.
	}
	\label{fig:pred_results}
  \end{figure}
  
We count computation costs (FLOPs) of each component in TransVG++ by setting input size of images as $640\times640$ and padded length of language expressions as 40, and report the statistic results in Table~\ref{tab:lvit_computation_cost}. Although the language branch accounts for the majority of parameters in TransVG++, the computation costs of vision branch (\emph{i.e.}, LViT) are much higher than that of language branch. Remarkably, in TransVG++ (base), the computation of LViT makes up 97.7\% of the total costs, which is 42.47 times higher than that of the language branch. Particularly, the computation costs introduced by our language adapters are negligible compared with the whole model. Take TransVG++ (tiny) as an example, the language adapters introduce 0.92G FLOPs, which is only 4.99\% of the language conditioned vision branch's computation costs, and only 3.64\% of the whole model's computation costs. This comparison verifies the computation cost efficiency of our proposed method.

One common concerned problem for to-date deep models is how to make them applicable to devices with limited resource (\emph{e.g.}, smartphone). As the language branch accounts for the major part of parameters, to slim our TransVG++, one potential direction is to replace BERT$_\text{BASE}$ with a smaller variety of BERT model (like TinyBERT~\cite{jiao2019tinybert}). Then, we can make use of network distillation techniques to decrease the performance gap between the slimmed model and the original TransVG++ model. To reduce the computation costs, two widely adopted strategies can be directly applied. One is the online token pruning technique~\cite{rao2021dynamicvit}, and the other is the sub-window attention strategy~\cite{chen2021simple,li2022exploring}. The above mentioned approaches are devoted to help improve model efficiency in each uni-modality. How to take multi-modal inputs to devise techniques to address the problem of model efficiency, including parameters efficiency and computation efficiency, is still not fully explored in the literature. We raise this potential direction as an open problem, and leave it for future investigation.

\begin{table}[t]
	\caption{Analysis on model size (number of parameters) of each component in our TransVG++ framework. TransVG++ (tiny/small/base) represent the models with ViT-tiny/small/base vision backbone networks, respectively. 
	}
	\footnotesize
	\begin{center}
	    \renewcommand\arraystretch{1.0}
		\scalebox{0.92}[0.92]{
			\setlength
			\tabcolsep{2.5pt}
			\begin{tabular}{l | c | c | c }
				\hline
				\toprule[1.2pt]
				Models & Vision Branch & Language Branch & Prediction Head \\
				\midrule[0.8pt]
				TransVG++ (tiny) & 6.39M (5.5\%) & 109.63M (94.4\%) & 0.12M (0.1\%)  \\
				TransVG++ (small) & 24.57M (18.3\%) & 109.78M (81.6\%) & 0.17M (0.1\%) \\
				TransVG++ (base)  & 96.33M (46.6\%) & 110.07M (53.3\%) & 0.26M (0.1\%) \\
				\bottomrule[1.2pt]
			\end{tabular}
		} 
	\end{center}
	\label{tab:lvit_model_size}
\end{table}

\begin{table}[t]
	\caption{Analysis on computation costs of each component in our TransVG++ framework in terms of FLOPs. TransVG++ (tiny/small/base) represent the models with ViT-tiny/small/base vision backbone networks, respectively. The FLOPs is obtained by setting the input image size as $640\times640$ and the padded query length as 40.}
	\footnotesize
	\begin{center}
	    \renewcommand\arraystretch{1.0}
		\scalebox{0.92}[0.92]{
			\setlength
			\tabcolsep{2.5pt}
			\begin{tabular}{l | c | c | c }
				\hline
				\toprule[1.2pt]
				Models & Vision Branch & Language Branch & Prediction Head \\
				\midrule[0.8pt]

				TransVG++ (tiny)  & 18.46G (73.1\%) & 6.80G (26.9\%)  & 232.46K (0.0\%) \\
				TransVG++ (small) & 72.86G (91.5\%) & 6.80G (8.5\%) & 330.76K (0.0\%) \\
				TransVG++ (base)  & 289.46G (97.7\%) & 6.80G (2.3\%) & 527.36K (0.0\%) \\

				\bottomrule[1.2pt]
			\end{tabular}
		} 
	\end{center}
	\label{tab:lvit_computation_cost}
\end{table}

\subsection{Qualitative Result}\label{sec4:examples}

We showcase the qualitative results of five examples from the RefCOCOg~\cite{mao2016generation} test set in Figure~\ref{fig:pred_results}. 
We observe that both our preliminary and advanced version framework can successfully model queries with complicated relationships, \emph{e.g.}, ``A woman see the sea water in back position'' in language expression (b). We depict both the predicted bounding box (with rectangles with blue line) and the \texttt{[REG]} token's attention over visual tokens of the last block. The visualization results empirically show that our frameworks generate interpretable attentions on the referred objct that corresponds to the object shape and location. Particularly, compared with the preliminary version, TransVG++ can generate more focus and clear attention, and thus can predict more correct coordinates given the referring expressions, as shown in example (c), (d) and (e).

\section{Conclusion}
In this paper, we present our TransVG (preliminary version) and TransVG++ (advanced version) frameworks to address the problem of visual grounding with Transformers. Instead of leveraging complex manually-designed fusion modules, TransVG uses a simple stack of Transformer encoders to perform the multi-modal fusion and reasoning for the visual grounding task. 
The advanced version TransVG++ takes a step further, upgrading to a purely Transformer-based architecture and removing the stand-alone fusion modules by integrating language referring information to the ViT backbone. Our TransVG++ serves as a simple, efficient, and accurate framework for visual grounding, and exhibits huge potential for future investigation.

\bibliographystyle{IEEEtran}
\bibliography{./reference}

\begin{thebibliography}{10}
\providecommand{\url}[1]{#1}
\csname url@samestyle\endcsname
\providecommand{\newblock}{\relax}
\providecommand{\bibinfo}[2]{#2}
\providecommand{\BIBentrySTDinterwordspacing}{\spaceskip=0pt\relax}
\providecommand{\BIBentryALTinterwordstretchfactor}{4}
\providecommand{\BIBentryALTinterwordspacing}{\spaceskip=\fontdimen2\font plus
\BIBentryALTinterwordstretchfactor\fontdimen3\font minus
  \fontdimen4\font\relax}
\providecommand{\BIBforeignlanguage}[2]{{%
\expandafter\ifx\csname l@#1\endcsname\relax
\typeout{** WARNING: IEEEtran.bst: No hyphenation pattern has been}%
\typeout{** loaded for the language `#1'. Using the pattern for}%
\typeout{** the default language instead.}%
\else
\language=\csname l@#1\endcsname
\fi
#2}}
\providecommand{\BIBdecl}{\relax}
\BIBdecl

\bibitem{nagaraja2016modeling}
V.~K. Nagaraja, V.~I. Morariu, and L.~S. Davis, ``Modeling context between
  objects for referring expression understanding,'' in \emph{Proceedings of the
  European Conference on Computer Vision (ECCV)}, 2016, pp. 792--807.

\bibitem{wang2019learning}
L.~Wang, Y.~Li, J.~Huang, and S.~Lazebnik, ``Learning two-branch neural
  networks for image-text matching tasks,'' \emph{IEEE Transactions on Pattern
  Analysis and Machine Intelligence (TPAMI)}, vol.~41, pp. 394--407, 2018.

\bibitem{hu2016natural}
R.~Hu, H.~Xu, M.~Rohrbach, J.~Feng, K.~Saenko, and T.~Darrell, ``Natural
  language object retrieval,'' in \emph{Proceedings of the IEEE Conference on
  Computer Vision and Pattern Recognition (CVPR)}, 2016, pp. 4555--4564.

\bibitem{li2017deep}
J.~Li, Y.~Wei, X.~Liang, F.~Zhao, J.~Li, T.~Xu, and J.~Feng, ``Deep
  attribute-preserving metric learning for natural language object retrieval,''
  in \emph{Proceedings of the 28th ACM International Conference on Multimedia
  (ACM MM)}.\hskip 1em plus 0.5em minus 0.4em\relax ACM, 2017, pp. 181--189.

\bibitem{chen2018real}
X.~Chen, L.~Ma, J.~Chen, Z.~Jie, W.~Liu, and J.~Luo, ``Real-time referring
  expression comprehension by single-stage grounding network,'' \emph{arXiv
  preprint arXiv:1812.03426}, 2018.

\bibitem{liao2020real}
Y.~Liao, S.~Liu, G.~Li, F.~Wang, Y.~Chen, C.~Qian, and B.~Li, ``A real-time
  cross-modality correlation filtering method for referring expression
  comprehension,'' in \emph{Proceedings of the IEEE Conference on Computer
  Vision and Pattern Recognition (CVPR)}, 2020, pp. 10\,880--10\,889.

\bibitem{yang2019fast}
Z.~Yang, B.~Gong, L.~Wang, W.~Huang, D.~Yu, and J.~Luo, ``A fast and accurate
  one-stage approach to visual grounding,'' in \emph{Proceedings of the IEEE
  International Conference on Computer Vision (ICCV)}, 2019, pp. 4683--4693.

\bibitem{dosovitskiy2020image}
A.~Dosovitskiy, L.~Beyer, A.~Kolesnikov, D.~Weissenborn, X.~Zhai,
  T.~Unterthiner, M.~Dehghani, M.~Minderer, G.~Heigold, S.~Gelly \emph{et~al.},
  ``An image is worth 16x16 words: Transformers for image recognition at
  scale,'' in \emph{International Conference on Learning Representations
  (ICLR)}, 2021.

\bibitem{wang2016learning}
L.~Wang, Y.~Li, and S.~Lazebnik, ``Learning deep structure-preserving
  image-text embeddings,'' in \emph{Proceedings of the IEEE Conference on
  Computer Vision and Pattern Recognition (CVPR)}, 2016, pp. 5005--5013.

\bibitem{liu2019learning}
D.~Liu, H.~Zhang, F.~Wu, and Z.-J. Zha, ``Learning to assemble neural module
  tree networks for visual grounding,'' in \emph{Proceedings of the IEEE
  International Conference on Computer Vision (ICCV)}, 2019, pp. 4673--4682.

\bibitem{yang2020improving}
Z.~Yang, T.~Chen, L.~Wang, and J.~Luo, ``Improving one-stage visual grounding
  by recursive sub-query construction,'' in \emph{Proceedings of the European
  Conference on Computer Vision (ECCV)}, 2020.

\bibitem{yu2018mattnet}
L.~Yu, Z.~Lin, X.~Shen, J.~Yang, X.~Lu, M.~Bansal, and T.~L. Berg, ``Mattnet:
  Modular attention network for referring expression comprehension,'' in
  \emph{Proceedings of the IEEE Conference on Computer Vision and Pattern
  Recognition (CVPR)}, 2018, pp. 1307--1315.

\bibitem{wang2019neighbourhood}
P.~Wang, Q.~Wu, J.~Cao, C.~Shen, L.~Gao, and A.~v.~d. Hengel, ``Neighbourhood
  watch: Referring expression comprehension via language-guided graph attention
  networks,'' in \emph{Proceedings of the IEEE Conference on Computer Vision
  and Pattern Recognition (CVPR)}, 2019, pp. 1960--1968.

\bibitem{yang2019dynamic}
S.~Yang, G.~Li, and Y.~Yu, ``Dynamic graph attention for referring expression
  comprehension,'' in \emph{Proceedings of the IEEE International Conference on
  Computer Vision (ICCV)}, 2019, pp. 4644--4653.

\bibitem{yang2020graph}
------, ``Graph-structured referring expression reasoning in the wild,'' in
  \emph{Proceedings of the IEEE Conference on Computer Vision and Pattern
  Recognition (CVPR)}, 2020, pp. 9952--9961.

\bibitem{hong2019learning}
R.~Hong, D.~Liu, X.~Mo, X.~He, and H.~Zhang, ``Learning to compose and reason
  with language tree structures for visual grounding,'' \emph{IEEE Transactions
  on Pattern Analysis and Machine Intelligence (TPAMI)}, 2019.

\bibitem{yu2016modeling}
L.~Yu, P.~Poirson, S.~Yang, A.~C. Berg, and T.~L. Berg, ``Modeling context in
  referring expressions,'' in \emph{Proceedings of the European Conference on
  Computer Vision (ECCV)}, 2016, pp. 69--85.

\bibitem{mao2016generation}
J.~Mao, J.~Huang, A.~Toshev, O.~Camburu, A.~L. Yuille, and K.~Murphy,
  ``Generation and comprehension of unambiguous object descriptions,'' in
  \emph{Proceedings of the IEEE Conference on Computer Vision and Pattern
  Recognition (CVPR)}, 2016, pp. 11--20.

\bibitem{kazemzadeh2014referitgame}
S.~Kazemzadeh, V.~Ordonez, M.~Matten, and T.~Berg, ``Referitgame: Referring to
  objects in photographs of natural scenes,'' in \emph{Conference on Empirical
  Methods in Natural Language Processing (EMNLP)}, 2014.

\bibitem{plummer2017flickr30k}
B.~A. Plummer, L.~Wang, C.~M. Cervantes, J.~C. Caicedo, J.~Hockenmaier, and
  S.~Lazebnik, ``Flickr30k entities: Collecting region-to-phrase
  correspondences for richer image-to-sentence models,'' \emph{International
  Journal of Computer Vision (IJCV)}, vol. 123, no.~1, p.~74, 2017.

\bibitem{deng2021transvg}
J.~Deng, Z.~Yang, T.~Chen, W.~Zhou, and H.~Li, ``Transvg: End-to-end visual
  grounding with transformers,'' in \emph{Proceedings of the IEEE International
  Conference on Computer Vision (ICCV)}, 2021, pp. 1769--1779.

\bibitem{hu2017modeling}
R.~Hu, M.~Rohrbach, J.~Andreas, T.~Darrell, and K.~Saenko, ``Modeling
  relationships in referential expressions with compositional modular
  networks,'' in \emph{Proceedings of the IEEE Conference on Computer Vision
  and Pattern Recognition (CVPR)}, 2017, pp. 1115--1124.

\bibitem{zhang2018grounding}
H.~Zhang, Y.~Niu, and S.-F. Chang, ``Grounding referring expressions in images
  by variational context,'' in \emph{Proceedings of the IEEE Conference on
  Computer Vision and Pattern Recognition (CVPR)}, 2018, pp. 4158--4166.

\bibitem{zhuang2018parallel}
B.~Zhuang, Q.~Wu, C.~Shen, I.~Reid, and A.~van~den Hengel, ``Parallel
  attention: A unified framework for visual object discovery through dialogs
  and queries,'' in \emph{Proceedings of the IEEE Conference on Computer Vision
  and Pattern Recognition (CVPR)}, 2018, pp. 4252--4261.

\bibitem{sadhu2019zero}
A.~Sadhu, K.~Chen, and R.~Nevatia, ``Zero-shot grounding of objects from
  natural language queries,'' in \emph{Proceedings of the IEEE International
  Conference on Computer Vision (ICCV)}, 2019, pp. 4694--4703.

\bibitem{girshick2014rich}
R.~Girshick, J.~Donahue, T.~Darrell, and J.~Malik, ``Rich feature hierarchies
  for accurate object detection and semantic segmentation,'' in
  \emph{Proceedings of the IEEE Conference on Computer Vision and Pattern
  Recognition (CVPR)}, 2014, pp. 580--587.

\bibitem{girshick2015fast}
R.~Girshick, ``Fast r-cnn,'' in \emph{Proceedings of the IEEE International
  Conference on Computer Vision (ICCV)}, 2015, pp. 1440--1448.

\bibitem{ren2016faster}
S.~Ren, K.~He, R.~Girshick, and J.~Sun, ``Faster r-cnn: Towards real-time
  object detection with region proposal networks.'' \emph{IEEE Transactions on
  Pattern Analysis and Machine Intelligence (TPAMI)}, vol.~39, no.~6, pp.
  1137--1149, 2016.

\bibitem{uijlings2013selective}
J.~R. Uijlings, K.~E. Van De~Sande, T.~Gevers, and A.~W. Smeulders, ``Selective
  search for object recognition,'' \emph{International Journal of Computer
  Vision (IJCV)}, vol. 104, pp. 154--171, 2013.

\bibitem{plummerCITE2018}
B.~A. Plummer, P.~Kordas, M.~H. Kiapour, S.~Zheng, R.~Piramuthu, and
  S.~Lazebnik, ``Conditional image-text embedding networks,'' in
  \emph{Proceedings of the European Conference on Computer Vision (ECCV)},
  2018, pp. 249--264.

\bibitem{anderson2018bottom}
P.~Anderson, X.~He, C.~Buehler, D.~Teney, M.~Johnson, S.~Gould, and L.~Zhang,
  ``Bottom-up and top-down attention for image captioning and visual question
  answering,'' in \emph{Proceedings of the IEEE conference on computer vision
  and pattern recognition (CVPR)}, 2018, pp. 6077--6086.

\bibitem{zhang2017discriminative}
Y.~Zhang, L.~Yuan, Y.~Guo, Z.~He, I.-A. Huang, and H.~Lee, ``Discriminative
  bimodal networks for visual localization and detection with natural language
  queries,'' in \emph{Proceedings of the IEEE Conference on Computer Vision and
  Pattern Recognition (CVPR)}, 2017, pp. 557--566.

\bibitem{yang2020relationship}
S.~Yang, G.~Li, and Y.~Yu, ``Relationship-embedded representation learning for
  grounding referring expressions,'' \emph{IEEE Transactions on Pattern
  Analysis and Machine Intelligence (TPAMI)}, vol.~43, no.~8, pp. 2765--2779,
  2020.

\bibitem{liu2019improving}
X.~Liu, Z.~Wang, J.~Shao, X.~Wang, and H.~Li, ``Improving referring expression
  grounding with cross-modal attention-guided erasing,'' in \emph{Proceedings
  of the IEEE Conference on Computer Vision and Pattern Recognition (CVPR)},
  2019, pp. 1950--1959.

\bibitem{bajaj2019g3raphground}
M.~Bajaj, L.~Wang, and L.~Sigal, ``G3raphground: Graph-based language
  grounding,'' in \emph{Proceedings of the IEEE International Conference on
  Computer Vision (ICCV)}, 2019, pp. 4281--4290.

\bibitem{chen2017query}
K.~Chen, R.~Kovvuri, and R.~Nevatia, ``Query-guided regression network with
  context policy for phrase grounding,'' in \emph{Proceedings of the IEEE
  International Conference on Computer Vision (ICCV)}, 2017, pp. 824--832.

\bibitem{dogan2019neural}
P.~Dogan, L.~Sigal, and M.~Gross, ``Neural sequential phrase grounding
  (seqground),'' in \emph{Proceedings of the IEEE Conference on Computer Vision
  and Pattern Recognition (CVPR)}, 2019, pp. 4175--4184.

\bibitem{redmon2018yolov3}
J.~Redmon and A.~Farhadi, ``Yolov3: An incremental improvement,''
  \emph{arXiv:1804.02767}, 2018.

\bibitem{bolme2010visual}
D.~S. Bolme, J.~R. Beveridge, B.~A. Draper, and Y.~M. Lui, ``Visual object
  tracking using adaptive correlation filters,'' in \emph{Proceedings of the
  IEEE Conference on Computer Vision and Pattern Recognition (CVPR)}, 2010, pp.
  2544--2550.

\bibitem{henriques2014high}
J.~F. Henriques, R.~Caseiro, P.~Martins, and J.~Batista, ``High-speed tracking
  with kernelized correlation filters,'' \emph{IEEE Transactions on Pattern
  Analysis and Machine Intelligence (TPAMI)}, vol.~37, pp. 583--596, 2014.

\bibitem{yang2020propagating}
S.~Yang, G.~Li, and Y.~Yu, ``Propagating over phrase relations for one-stage
  visual grounding,'' in \emph{Proceedings of the European Conference on
  Computer Vision (ECCV)}.\hskip 1em plus 0.5em minus 0.4em\relax Springer,
  2020, pp. 589--605.

\bibitem{vaswani2017attention}
A.~Vaswani, N.~Shazeer, N.~Parmar, J.~Uszkoreit, L.~Jones, A.~N. Gomez,
  L.~Kaiser, and I.~Polosukhin, ``Attention is all you need,'' in
  \emph{Advances in Neural Information Processing Systems (NeurIPS)}, 2017.

\bibitem{mikolov2010recurrent}
T.~Mikolov, M.~Karafi{\'a}t, L.~Burget, J.~{\v{C}}ernock{\`y}, and
  S.~Khudanpur, ``Recurrent neural network based language model,'' in
  \emph{InterSpeech}, 2010.

\bibitem{tai2015improved}
K.~S. Tai, R.~Socher, and C.~D. Manning, ``Improved semantic representations
  from tree-structured long short-term memory networks,''
  \emph{arXiv:1503.00075}, 2015.

\bibitem{hochreiter1997long}
S.~Hochreiter and J.~Schmidhuber, ``Long short-term memory,'' \emph{Neural
  computation}, vol.~9, pp. 1735--1780, 1997.

\bibitem{carion2020end}
N.~Carion, F.~Massa, G.~Synnaeve, N.~Usunier, A.~Kirillov, and S.~Zagoruyko,
  ``End-to-end object detection with transformers,'' in \emph{Proceedings of
  the European Conference on Computer Vision (ECCV)}, 2020, pp. 213--229.

\bibitem{chen2020generative}
M.~Chen, A.~Radford, R.~Child, J.~Wu, H.~Jun, D.~Luan, and I.~Sutskever,
  ``Generative pretraining from pixels,'' in \emph{International Conference on
  Machine Learning (ICML)}, 2020, pp. 1691--1703.

\bibitem{yang2020learning}
F.~Yang, H.~Yang, J.~Fu, H.~Lu, and B.~Guo, ``Learning texture transformer
  network for image super-resolution,'' in \emph{Proceedings of the IEEE
  Conference on Computer Vision and Pattern Recognition (CVPR)}, 2020, pp.
  5791--5800.

\bibitem{zeng2020learning}
Y.~Zeng, J.~Fu, and H.~Chao, ``Learning joint spatial-temporal transformations
  for video inpainting,'' in \emph{Proceedings of the European Conference on
  Computer Vision (ECCV)}, 2020, pp. 528--543.

\bibitem{touvron2021training}
H.~Touvron, M.~Cord, M.~Douze, F.~Massa, A.~Sablayrolles, and H.~J{\'e}gou,
  ``Training data-efficient image transformers \& distillation through
  attention,'' in \emph{International Conference on Machine Learning (ICML)},
  2021, pp. 10\,347--10\,357.

\bibitem{liu2021swin}
Z.~Liu, Y.~Lin, Y.~Cao, H.~Hu, Y.~Wei, Z.~Zhang, S.~Lin, and B.~Guo, ``Swin
  transformer: Hierarchical vision transformer using shifted windows,'' in
  \emph{Proceedings of the IEEE International Conference on Computer Vision
  (ICCV)}, 2021, pp. 10\,012--10\,022.

\bibitem{meng2021conditional}
D.~Meng, X.~Chen, Z.~Fan, G.~Zeng, H.~Li, Y.~Yuan, L.~Sun, and J.~Wang,
  ``Conditional detr for fast training convergence,'' in \emph{Proceedings of
  the IEEE International Conference on Computer Vision (ICCV)}, 2021, pp.
  3651--3660.

\bibitem{chen2021simple}
W.~Chen, X.~Du, F.~Yang, L.~Beyer, X.~Zhai, T.-Y. Lin, H.~Chen, J.~Li, X.~Song,
  Z.~Wang \emph{et~al.}, ``A simple single-scale vision transformer for object
  localization and instance segmentation,'' \emph{arXiv preprint
  arXiv:2112.09747}, 2021.

\bibitem{li2022exploring}
Y.~Li, H.~Mao, R.~Girshick, and K.~He, ``Exploring plain vision transformer
  backbones for object detection,'' \emph{arXiv preprint arXiv:2203.16527},
  2022.

\bibitem{devlin2018bert}
J.~Devlin, M.-W. Chang, K.~Lee, and K.~Toutanova, ``Bert: Pre-training of deep
  bidirectional transformers for language understanding,''
  \emph{arXiv:1810.04805}, 2018.

\bibitem{chen2020uniter}
Y.-C. Chen, L.~Li, L.~Yu, A.~El~Kholy, F.~Ahmed, Z.~Gan, Y.~Cheng, and J.~Liu,
  ``Uniter: Universal image-text representation learning,'' in
  \emph{Proceedings of the European Conference on Computer Vision (ECCV)},
  2020, pp. 104--120.

\bibitem{li2020oscar}
X.~Li, X.~Yin, C.~Li, P.~Zhang, X.~Hu, L.~Zhang, L.~Wang, H.~Hu, L.~Dong,
  F.~Wei \emph{et~al.}, ``Oscar: Object-semantics aligned pre-training for
  vision-language tasks,'' in \emph{Proceedings of the European Conference on
  Computer Vision (ECCV)}, 2020, pp. 121--137.

\bibitem{lu2019vilbert}
J.~Lu, D.~Batra, D.~Parikh, and S.~Lee, ``Vilbert: Pretraining task-agnostic
  visiolinguistic representations for vision-and-language tasks,'' in
  \emph{Advances in Neural Information Processing Systems (NeurIPS)}, 2019.

\bibitem{su2019vl}
W.~Su, X.~Zhu, Y.~Cao, B.~Li, L.~Lu, F.~Wei, and J.~Dai, ``Vl-bert:
  Pre-training of generic visual-linguistic representations,'' \emph{arXiv
  preprint arXiv:1908.08530}, 2019.

\bibitem{yang2020tap}
Z.~Yang, Y.~Lu, J.~Wang, X.~Yin, D.~Florencio, L.~Wang, C.~Zhang, L.~Zhang, and
  J.~Luo, ``Tap: Text-aware pre-training for text-vqa and text-caption,''
  \emph{arXiv:2012.04638}, 2020.

\bibitem{huang2020pixel}
Z.~Huang, Z.~Zeng, B.~Liu, D.~Fu, and J.~Fu, ``Pixel-bert: Aligning image
  pixels with text by deep multi-modal transformers,'' \emph{arXiv preprint
  arXiv:2004.00849}, 2020.

\bibitem{kim2021vilt}
W.~Kim, B.~Son, and I.~Kim, ``Vilt: Vision-and-language transformer without
  convolution or region supervision,'' in \emph{International Conference on
  Machine Learning (ICML)}, 2021.

\bibitem{li2021align}
J.~Li, R.~R. Selvaraju, A.~D. Gotmare, S.~Joty, C.~Xiong, and S.~Hoi, ``Align
  before fuse: Vision and language representation learning with momentum
  distillation,'' in \emph{Advances in Neural Information Processing Systems
  (NeurIPS)}, 2021.

\bibitem{wang2021simvlm}
Z.~Wang, J.~Yu, A.~W. Yu, Z.~Dai, Y.~Tsvetkov, and Y.~Cao, ``Simvlm: Simple
  visual language model pretraining with weak supervision,'' in
  \emph{International Conference on Learning Representations (ICLR)}, 2022.

\bibitem{kamath2021mdetr}
A.~Kamath, M.~Singh, Y.~LeCun, G.~Synnaeve, I.~Misra, and N.~Carion,
  ``Mdetr-modulated detection for end-to-end multi-modal understanding,'' in
  \emph{Proceedings of the IEEE International Conference on Computer Vision
  (ICCV)}, 2021, pp. 1780--1790.

\bibitem{ba2016layer}
J.~L. Ba, J.~R. Kiros, and G.~E. Hinton, ``Layer normalization,''
  \emph{arXiv:1607.06450}, 2016.

\bibitem{he2016deep}
K.~He, X.~Zhang, S.~Ren, and J.~Sun, ``Deep residual learning for image
  recognition,'' in \emph{Proceedings of the IEEE Conference on Computer Vision
  and Pattern Recognition (CVPR)}, 2016, pp. 770--778.

\bibitem{parmar2018image}
N.~Parmar, A.~Vaswani, J.~Uszkoreit, L.~Kaiser, N.~Shazeer, A.~Ku, and D.~Tran,
  ``Image transformer,'' in \emph{International Conference on Machine Learning
  (ICML)}, 2018, pp. 4055--4064.

\bibitem{dehghani2018universal}
M.~Dehghani, S.~Gouws, O.~Vinyals, J.~Uszkoreit, and L.~Kaiser, ``Universal
  transformers,'' in \emph{International Conference on Learning Representations
  (ICLR)}, 2018.

\bibitem{raffel2019exploring}
C.~Raffel, N.~Shazeer, A.~Roberts, K.~Lee, S.~Narang, M.~Matena, Y.~Zhou,
  W.~Li, and P.~J. Liu, ``Exploring the limits of transfer learning with a
  unified text-to-text transformer,'' \emph{arXiv:1910.10683}, 2019.

\bibitem{liu2021gpt}
X.~Liu, Y.~Zheng, Z.~Du, M.~Ding, Y.~Qian, Z.~Yang, and J.~Tang, ``Gpt
  understands, too,'' \emph{arXiv preprint arXiv:2103.10385}, 2021.

\bibitem{houlsby2019parameter}
N.~Houlsby, A.~Giurgiu, S.~Jastrzebski, B.~Morrone, Q.~De~Laroussilhe,
  A.~Gesmundo, M.~Attariyan, and S.~Gelly, ``Parameter-efficient transfer
  learning for nlp,'' in \emph{International Conference on Machine Learning
  (ICML)}, 2019, pp. 2790--2799.

\bibitem{pfeiffer2020adapterfusion}
J.~Pfeiffer, A.~Kamath, A.~R{\"u}ckl{\'e}, K.~Cho, and I.~Gurevych,
  ``Adapterfusion: Non-destructive task composition for transfer learning,''
  \emph{arXiv preprint arXiv:2005.00247}, 2020.

\bibitem{rezatofighi2019generalized}
H.~Rezatofighi, N.~Tsoi, J.~Gwak, A.~Sadeghian, I.~Reid, and S.~Savarese,
  ``Generalized intersection over union: A metric and a loss for bounding box
  regression,'' in \emph{Proceedings of the IEEE Conference on Computer Vision
  and Pattern Recognition (CVPR)}, 2019, pp. 658--666.

\bibitem{escalante2010segmented}
H.~J. Escalante, C.~A. Hern{\'a}ndez, J.~A. Gonzalez, A.~L{\'o}pez-L{\'o}pez,
  M.~Montes, E.~F. Morales, L.~E. Sucar, L.~Villase{\~n}or, and M.~Grubinger,
  ``The segmented and annotated iapr tc-12 benchmark,'' \emph{Computer Vision
  and Image Understanding (CVIU)}, vol. 114, pp. 419--428, 2010.

\bibitem{young2014image}
P.~Young, A.~Lai, M.~Hodosh, and J.~Hockenmaier, ``From image descriptions to
  visual denotations: New similarity metrics for semantic inference over event
  descriptions,'' \emph{Annual Meeting of the Association for Computational
  Linguistics (ACL)}, vol.~2, pp. 67--78, 2014.

\bibitem{li2021referring}
M.~Li and L.~Sigal, ``Referring transformer: A one-step approach to multi-task
  visual grounding,'' \emph{Advances in Neural Information Processing Systems
  (NeurIPS)}, vol.~34, 2021.

\bibitem{liu2020learning}
Y.~Liu, B.~Wan, X.~Zhu, and X.~He, ``Learning cross-modal context graph for
  visual grounding,'' in \emph{Proceedings of the AAAI Conference on Artificial
  Intelligence (AAAI)}, vol.~34, no.~07, 2020, pp. 11\,645--11\,652.

\bibitem{lin2014microsoft}
T.-Y. Lin, M.~Maire, S.~Belongie, J.~Hays, P.~Perona, D.~Ramanan,
  P.~Doll{\'a}r, and C.~L. Zitnick, ``Microsoft coco: Common objects in
  context,'' in \emph{Proceedings of the European Conference on Computer Vision
  (ECCV)}, 2014, pp. 740--755.

\bibitem{he2017mask}
K.~He, G.~Gkioxari, P.~Doll{\'a}r, and R.~Girshick, ``Mask r-cnn,'' in
  \emph{Proceedings of the IEEE International Conference on Computer Vision
  (ICCV)}, 2017, pp. 2961--2969.

\bibitem{du2022vgtr}
Y.~Du, Z.~Fu, Q.~Liu, and Y.~Wang, ``Visual grounding with transformers,'' in
  \emph{Proceedings of the IEEE International Conference on Multimedia \& Expo
  (ICME)}, 2022.

\bibitem{chen2014semantic}
L.-C. Chen, G.~Papandreou, I.~Kokkinos, K.~Murphy, and A.~L. Yuille, ``Semantic
  image segmentation with deep convolutional nets and fully connected crfs,''
  \emph{arXiv preprint arXiv:1412.7062}, 2014.

\bibitem{kovvuri2018pirc}
R.~Kovvuri and R.~Nevatia, ``Pirc net: Using proposal indexing, relationships
  and context for phrase grounding,'' in \emph{Proceedings of the Asia
  Conference on Computer Vision (ACCV)}, 2018, pp. 451--467.

\bibitem{yu2018rethinking}
Z.~Yu, J.~Yu, C.~Xiang, Z.~Zhao, Q.~Tian, and D.~Tao, ``Rethinking diversified
  and discriminative proposal generation for visual grounding,'' in
  \emph{International Joint Conference on Artificial Intelligence (IJCAI)},
  2018.

\bibitem{jiao2019tinybert}
X.~Jiao, Y.~Yin, L.~Shang, X.~Jiang, X.~Chen, L.~Li, F.~Wang, and Q.~Liu,
  ``Tinybert: Distilling bert for natural language understanding,'' \emph{arXiv
  preprint arXiv:1909.10351}, 2019.

\bibitem{rao2021dynamicvit}
Y.~Rao, W.~Zhao, B.~Liu, J.~Lu, J.~Zhou, and C.-J. Hsieh, ``Dynamicvit:
  Efficient vision transformers with dynamic token sparsification,''
  \emph{Advances in neural information processing systems (NeurIPS)}, vol.~34,
  2021.

\end{thebibliography}

\end{document}